\documentclass[acmlarge]{acmart}
\usepackage{amsfonts,subfigure,enumitem,multirow}
\usepackage{algorithmic}
\usepackage{graphicx}
\usepackage[ruled,vlined]{algorithm2e}
\usepackage{graphicx}
\usepackage{subcaption}
\usepackage{xcolor}
\usepackage{color}
\usepackage{makecell}

\AtBeginDocument{%
  \providecommand\BibTeX{{%
    \normalfont B\kern-0.5em{\scshape i\kern-0.25em b}\kern-0.8em\TeX}}}

\setcopyright{acmcopyright}
\copyrightyear{2018}
\acmYear{2018}
\acmDOI{10.1145/1122445.1122456}

\acmConference[Woodstock '18]{Woodstock '18: ACM Symposium on Neural
  Gaze Detection}{June 03--05, 2018}{Woodstock, NY}
\acmBooktitle{Woodstock '18: ACM Symposium on Neural Gaze Detection,
  June 03--05, 2018, Woodstock, NY}
\acmPrice{15.00}
\acmISBN{978-1-4503-XXXX-X/18/06}

\begin{document}

\title[STContext]{STContext: A Multifaceted Dataset for Developing Context-aware Spatio-temporal Crowd Mobility Prediction Models}

\author{Liyue Chen}
\authornote{Both authors contributed equally to this work.}
\email{chenliyue2019@gmail.com}
\author{Jiangyi Fang}
\authornotemark[1]
\affiliation{%
  \institution{Peking University}
  % \streetaddress{P.O. Box 1212}
  \city{Beijing}
  \country{China}
}

\author{Tengfei Liu}
\affiliation{%
  \institution{Peking University}
  % \city{Rocquencourt}
  \country{China}
}

\author{Fangyuan Gao}
\affiliation{%
  \institution{Peking University}
  % \city{Rocquencourt}
 \country{China}
}

\author{Leye Wang}
\authornote{Corresponding author.}
\affiliation{%
  \institution{Peking University}
  % \city{Rocquencourt}
  \country{China}
}

\renewcommand{\shortauthors}{Trovato and Tobin, et al.}

\begin{abstract}
In smart cities, context-aware spatio-temporal crowd flow prediction (STCFP) models leverage contextual features (e.g., weather) to identify unusual crowd mobility patterns and enhance prediction accuracy. However, the best practice for incorporating contextual features remains unclear due to inconsistent usage of contextual features in different papers. Developing a multifaceted dataset with rich types of contextual features and STCFP scenarios is crucial for establishing a principled context modeling paradigm. Existing open crowd flow datasets lack an adequate range of contextual features, which poses an urgent requirement to build a multifaceted dataset to fill these research gaps.
To this end, we create \texttt{STContext}, a multifaceted dataset for developing context-aware STCFP models. Specifically, \texttt{STContext} provides nine spatio-temporal datasets across five STCFP scenarios and includes ten contextual features, including weather, air quality index, holidays, points of interest, road networks, etc.
Besides, we propose a unified workflow for incorporating contextual features into deep STCFP methods, with steps including feature transformation, dependency modeling, representation fusion, and training strategies. Through extensive experiments, we have obtained several useful guidelines for effective context modeling and insights for future research. The \texttt{STContext} is open-sourced at \textbf{\url{https://github.com/Liyue-Chen/STContext}}.
\end{abstract}




\maketitle

\section{Introduction}

With rapid urbanization, ubiquitous smart devices are collecting massive data with timestamps and location information. Accurately predicting these spatio-temporal (ST) data is the basis for enterprises and governments to make informed decisions for many real-world applications, like bike-sharing \cite{li_traffic_2015,yang_mobility_2016,chai_multi_graph_2018, LiBikeTKDE2019} and ride-hailing \cite{tong_simpler_2017,ke_short-term_2017,wang_deepsd_2017, Saadallah_taxi_2020}. 
In the realm of ST prediction, contextual features (e.g., weather) have proven to be beneficial in a wide variety of applications for distinguishing unusual crowd mobility patterns \cite{surban_imwut_2023,hoang_fccf:_2016,zhu_deep_2017,zhang2017deep,yao2018DeepMulti}. For example, heavy rains and strong winds may decrease the utilization of bike-sharing and online ride-hailing services \cite{hoang_fccf:_2016, LiBikeTKDE2019}.

\begin{table}
  \small
  \caption{Comparing \texttt{STContext} to widely-used open-source crowd flow datasets, we examine studies employing these datasets. `AQI' refers to the air quality index, `POI' stands for point of interest, and `A.D.' denotes administrative division. `TP' and `SP' are temporal and spatial position features (e.g., hour of day and geographical coordinates).}
  \label{tab: dataset_comparison}
  \centering
  % \resizebox{.95\textwidth}{!}{
  \setlength{\tabcolsep}{1mm}{
  \begin{tabular}{lcccccccccc}
\toprule
\multirow{2}{*}{\textbf{Dataset}} & \multicolumn{2}{c}{\textbf{Weather}} & \multirow{2}{*}{\textbf{AQI}} & \multirow{2}{*}{\textbf{Holiday}} & \multirow{2}{*}{\textbf{POI}} & \multirow{2}{*}{\textbf{Road}} & \multirow{2}{*}{\textbf{Demographics}} & \multirow{2}{*}{\textbf{A.D.}} & \multirow{2}{*}{\textbf{TP}} & \multirow{2}{*}{\textbf{SP}}  \\
\cmidrule(lr){2-3} 
 &  \textit{Historical} & \textit{Forecast} &  & & &  &  &  \\
\midrule
METR-LA~\cite{li2017diffusion,STGCN_2018,graphwavenet_2019,STAEformer_2023,LightCTS_2023} & - & - & - & - & - & - & - & - & - & $\checkmark$ \\
Loop Seattle~\cite{cui2017deep}  & - & - & - & - & - & - & - & - & - & -\\
Q-Traffic~\cite{liao_2018,traffic_prediction_methods_tits_2022}  & - & - & -  & $\checkmark$ & - & $\checkmark$ & - & - & - & - \\
PEMS~\cite{li2017diffusion,STGCN_2018, ASTGCN_2019, song2020spatial,decoupled_STG_2024} & - & - & -  & - & - & - & - & - & - & -\\
Los-loop, SZ-taxi~\cite{TGCN_2020} & - & - & - & - & - & - & - & - & - & -\\
LargeST~\cite{liu2023largest} & - & - & - & - & - & $\checkmark$ & - & - & - & $\checkmark$ \\
NYC-Risk, CHI-Risk~\cite{GSNet_2021} & $\checkmark$ & -& -  & - & $\checkmark$ & $\checkmark$ & - & - & - & -\\
Beijing subway~\cite{multigraph_subway_flow_2020,zhang_2021_metro} & $\checkmark$ & - & $\checkmark$ & - & -  & - & - & - & - 
& $\checkmark$ \\
SHMetro, HZMetro~\cite{liu_inter_metro_2022} & - & - & - & - & -  & - & - & - & - & $\checkmark$\\
TaxiBJ~\cite{zhang2017deep, urbanfm_2019, fine_grained_2021} & $\checkmark$ & - & - & $\checkmark$ & - & - & - & - & - & - \\
NYCTaxi1401~\cite{ACFM_2018} & $\checkmark$ & - & - & $\checkmark$ & - & - & - & - & - & -\\
NYCTaxi1601, NYCBike1608~\cite{lin_dsan_2020} & - & - & - & - & - & - & - & - & $\checkmark$ & -\\
NYCTaxi~\cite{liu2019contextualized,yao_revisit_2019} & $\checkmark$ & - & - & - & - & - & - & - & - & -\\
UUKG~\cite{ning2023uukg} & - & - & - & - & $\checkmark$ & $\checkmark$ & - & $\checkmark$ & - & $\checkmark$ \\
\midrule
\textbf{\texttt{STContext} (Ours)} & $\checkmark$ & $\checkmark$ & $\checkmark$ & $\checkmark$ & $\checkmark$ & $\checkmark$ & $\checkmark$ & $\checkmark$ & $\checkmark$ & $\checkmark$  \\
\bottomrule
  \end{tabular}}
  \vspace{-1em}
\end{table}

Pioneering research has made a great effort in developing context-aware spatio-temporal crowd flow prediction (STCFP) models to enhance accuracy \cite{li_traffic_2015, zhang2017deep, lin2019deepstn+, fine_grained_2021,ning2023uukg}. Although these works improve predictions in specific applications, they utilize different contextual features and modeling designs, resulting in inconsistent and incomparable results. For example, as illustrated in Table \ref{tab: dataset_comparison}, the incorporated contextual features differ among papers and datasets. The best practice for incorporating contextual features into STCFP remains unclear. 
Therefore, there is an urgent need to develop a principled context modeling paradigm and a multifaceted dataset that includes rich types of contextual features and STCFP scenarios to systematically guide contextual feature processing, modeling, and fusion methods. Such a dataset would enable a fair comparison of existing techniques and features, provide valuable insights for developing advanced contextual modeling techniques, and comprehensively evaluate the ability of context-aware STCFP models.

However, existing widely-used open crowd flow datasets are not sufficiently qualified due to their limited consideration of contextual features. As shown in Table~\ref{tab: dataset_comparison}, most datasets (e.g., PEMS \cite{li2017diffusion} and LargeST \cite{liu2023largest}) cover only two or three types of contextual features. Furthermore, some valuable contextual features have been disregarded. For instance, while previous studies have emphasized the importance of forecasted weather features for STCFP \cite{deep_fusion_net_2018}, there is currently no openly accessible dataset that includes forecasted weather information. Hence, there is an urgent need for a comprehensive and multifaceted context dataset, while building such a dataset is non-trivial and will encounter the following challenges.

First, \textit{\textbf{what types of context should we gather?}} As there are numerous contextual features, the determination of the range of features to collect is unclear. Second, \textbf{\textit{how can we create a taxonomy of these contextual features?}} Since context modeling techniques may be designed based on context characteristics (e.g., using RNNs to model temporal dependency of weather \cite{ke_short-term_2017}), developing such a taxonomy would not only help discover the similarity among contextual features but also facilitate the context modeling process. Last, \textbf{\textit{where should we collect context data?}} There may be multiple data sources available for a particular contextual feature. For instance, as presented in Table \ref{tab: data_source}, there are at least five different data sources for historical weather information. With such an abundance of options, how to choose suitable sources needs careful considerations.

Taking the above challenges into account, we propose \texttt{STContext}, a comprehensive dataset for developing context-aware STCFP models. With \texttt{STContext}, we are then able to conduct experiments with a rich set of contextual features and provide a preliminary analysis of how these contextual features may impact STCFP performance. In summary, our main contributions include:

\begin{itemize}[leftmargin=1em]
    \item We create a multifaceted dataset that provides valuable contextual data, including weather, air quality index, holidays, temporal position (e.g., hour of day), points of interest, road network, administrative division, demographic data, and spatial position (e.g., geographical coordinates) across five STCFP tasks. To our knowledge, this is the most comprehensive contextual dataset for developing context-aware STCFP models up to date.

    \item We investigate contextual features in recent research from over ten reputable venues (UbiComp, TMC, KDD, etc.) and identify ten commonly used and publicly accessible contextual features (POI, weather, road, etc.). Additionally, we extensively investigate publicly accessible context data sources that not only support this study but also may benefit a spectrum of researchers interested in STCFP context research. Furthermore, we categorize identified contextual features into three types: spatial, temporal, and spatio-temporal, based on their variations in time and space dimensions. This taxonomy can help design appropriate strategies for incorporating different contextual features.
    
    \item We present a unified workflow paradigm for incorporating contextual features into deep STCFP models. With this paradigm, we evaluate existing representative context modeling techniques on \texttt{STContext}, gaining valuable insights for future research. Furthermore, we offer open-source codes and instructions for utilizing our dataset and replicating our experiments. We hope \texttt{STContext} can provide an opportunity for the STCFP research community to develop new and effectively generalizable context modeling techniques.
\end{itemize}

\section{Related Work}

\subsection{Spatio-temporal Prediction Datasets} 

Just as ImageNet has significantly advanced computer vision research~\cite{imageNet_2009}, spatio-temporal datasets from traffic sensors, GPS trajectories, and remote sensing have sparked breakthroughs in urban computing, traffic management, and intelligent transportation systems.
For instance, crowd flow datasets like METR-LA and PeMS have been instrumental in advancing deep learning techniques for traffic prediction \cite{li2017diffusion,ASTGCN_2019,song2020spatial}. Similarly, datasets like TaxiBJ and BikeNYC have enabled researchers to explore the patterns and anomalies in urban mobility \cite{zhang2017deep}. 
Q-Traffic provides publicly available datasets that include diverse auxiliary information, such as road structure and public holidays \cite{liao_2018}.
Recently, LargeST provided a large-scale benchmark with 8,600 sensors across California, offering a more realistic scale representation of traffic networks compared to existing datasets \cite{liu2023largest}.

Although existing spatio-temporal crowd flow datasets have made significant progress in data scale and task diversity, they lack comprehensive contextual features. As shown in Table~\ref{tab: dataset_comparison}, even though open datasets like LargeST~\cite{liu2023largest} and UUKG~\cite{ning2023uukg} include POI and road data, they still are missing other contextual features such as weather. Moreover, most datasets lack any contextual features, highlighting the urgent need for a multifaceted dataset with rich context. In contrast, \texttt{STContext} offers a comprehensive collection of contextual data, including weather conditions, holidays, POI, AQI, road network details, and demographic data across five STCFP scenarios. To our knowledge, it is the most extensive dataset for developing context-aware STCFP models to date.

\subsection{Context-aware Spatio-temporal Prediction Models} 
Recent advances in deep learning have significantly improved spatio-temporal crowd flow prediction. From the temporal perspective, various neural network architectures have been introduced to capture dependencies, including RNNs like LSTM \cite{hochreiter1997long} and GRU \cite{chung2014empirical}, as well as temporal convolutional networks like WaveNet \cite{WaveNet_2016} and TCN \cite{gated_tcn_2017}. More recently, attention-based Transformer models, such as Informer \cite{informer_2021}, FEDformer \cite{FEDformer_2022}, and Non-stationary Transformer \cite{liu2022non}, have emerged for time series prediction. Spatially, cities can be divided into grids, with CNNs used to extract features from nearby areas \cite{ke_short-term_2017,curbGan_2020,zhang_daily_2021}. Stacking multiple convolution layers allows for capturing distant dependencies \cite{zhang2017deep, zhang_flow_2019}. Graphs offer a more flexible representation of spatial correlations, with GNNs like GCN \cite{grarep_2015,chai_multi_graph_2018,STGCN_2018,graphwavenet_2019} and GAT \cite{ASTGCN_2019, GMAN_AAAI2020} widely used for modeling spatial dependencies.

These works propose effective spatio-temporal models for handling diverse types of crowd flow data, achieving promising results. Notably, these studies design various techniques to effectively model context. For example, \textit{ST-ResNet} \cite{zhang2017deep} employs embedding layers to learn context representations, which are then fused with the crowd flow representation.
Inspired by the idea that context may influence crowds like a switch, \textit{gating} mechanisms map context into scaling factors that adjust the crowd flow representation \cite{zhang_flow_2019}.
However, the lack of a comprehensive evaluation benchmark limits the generalizability of these context-aware modeling techniques, even if they perform well on specific datasets, as recent research has shown \cite{context_generalizability}. While prior studies have developed benchmarks for evaluating spatio-temporal models, they typically exclude contextual features \cite{dl_traffic_2021, dynamic_graph_benchmark_2023}. This paper focuses on building a comprehensive context dataset for thoroughly evaluating existing context-aware STCFP models and helping develop more advanced modeling techniques.

\section{The \texttt{STContext} Dataset Construction}

\subsection{Selection of Contextual Features} \label{sec: context_selection}
To determine which types of context should be collected, we conducted a comprehensive review of spatio-temporal prediction papers presented at over ten renowned venues, including IEEE TMC, UbiComp, KDD, ICDE, WWW, WSDM, CIKM, NeurIPS, AAAI, and IJCAI.
From over 500 spatio-temporal prediction papers, we identified more than 80 papers that used contextual features to improve crowd mobility prediction accuracy.
Based on this literature review, with attention to the public availability of contextual data, we select ten vital types of contextual features: historical weather, weather forecast, air quality index (AQI), holidays, temporal position, points of interest (POI), road network information, demographic data, administrative division, and spatial position. Table~\ref{tab: features_investigate} lists representative studies and the contextual features they utilized.

\begin{table*}[ht!]
	\footnotesize
	\caption{Contextual features in previous STCFP literature. `TP' and `SP' are temporal and spatial position features (e.g., hour of day and geographical coordinates), respectively. (AQI: Air Quality Index; T: Temperature; H: Humidity; V: Visibility; WS: Wind Speed; WD: Wind Degree; S: Weather State)}
    \vspace{-.5em}
	\label{tab: features_investigate}
	\begin{center}
		\resizebox{0.97\textwidth}{!}{
			\begin{tabular}{c|c|c|c|c|ccccc}
\hline
\textbf{Prediction Task}& \textbf{Literature} & \textbf{Spatio-temporal Context} & \textbf{Temporal Context} & \textbf{Spatial Context} & \textbf{Venue} \\

\hline
\multirow{4}{*}{Bike-sharing} & \emph{Li et al.} \cite{li_traffic_2015} & Historical Weather (T;WS;S) & - & - & SIGSPATIAL '15\\ 
& \emph{Yang et al.} \cite{yang_mobility_2016} & Historical Weather (T;H;V;WS;S) & TP & - & MobiSys '16 \\ 
& \emph{Li et al.} \cite{LiBikeTKDE2019} & Historical Weather (T;WS;S) & Holiday, TP & - & TKDE '19 \\ 
& \emph{He et al.} \cite{he_escooter_2022} & Historical Weather (T;H;WS;S) & - & POI, Demographic & WWW '21 \\ 

\hline

\multirow{5}{*}{Ride-sharing} & \emph{Tong et al.} \cite{tong_simpler_2017} & Historical Weather (T;H;WS;WD;S), AQI & Holiday, TP & POI & KDD '17  \\ 
& \emph{Ke et al.} \cite{ke_short-term_2017} & Historical Weather (T;H;V;WS;S) & TP & - & TR Part C '17 \\ 
& \emph{Wang et al.} \cite{wang_deepsd_2017} & Historical Weather (T;S), AQI & TP & SP & ICDE '17 \\ 
& \emph{Yao et al.} \cite{yao2018DeepMulti} & Historical Weather (T;S) & Holiday & - & AAAI '17 \\ 
& \emph{Saadallah et al.} \cite{Saadallah_taxi_2020} & Historical Weather (T;WS;S) & - & POI & TKDE '18 \\ 

\hline
\multirow{3}{*}{Metro Passenger Flow} & \emph{Liu et al.} \cite{liu_deeppf:_2019} & Historical Weather (S) & TP & - & TR Part C '19 \\ 
& \emph{Wang et al.} \cite{wang_subways_two_way_2022} & - & - & POI, Demographic & TITS '22 \\ 
& \emph{Xu et al.} \cite{adaptive_fusion_metro_2023} & Historical Weather (T;H;WS;V), AQI & Holiday, TP & - & TITS '23 \\ 

\hline
\multirow{7}{*}{Traffic Flow} & \emph{Barnes et al.} \cite{BarnesBCFTX20} & - & TP & - & KDD '20 \\ 
& \emph{Zheng et al} \cite{GMAN_AAAI2020} & - & TP & Road & AAAI '20 \\ 
& \emph{Pan et al.} \cite{meta_learning_traffic_2019} & - & - & POI, Road & KDD '19 \\ 
& \emph{Zhang et al.} \cite{traffic_diffusion_2021} & Historical Weather (WS;T;S) & Holiday, TP & - & AAAI '21 \\ 
& \emph{Yuan et al.} \cite{yuan_demand_2021} & Historical Weather (S) & Holiday, TP & - & ICDE '21 \\ 
& \emph{Kim et al.} \cite{traffic_adversarial_2022} & Historical Weather (T;S) & TP & - & ICDE '22 \\ 
& \emph{Han et al.} \cite{han_traffic_activity_2023} & - & -& Road, SP & CIKM '23 \\ 

\hline
\multirow{11}{*}{Crowd Flow} & \emph{Zhang et al.} \cite{zhang2017deep} & Historical Weather (T;WS;S) & Holiday & - & AAAI '17 \\ 
& \emph{Lin et al.} \cite{lin2019deepstn+} & - & TP & POI & AAAI '19 \\ 
& \emph{Sun et al.} \cite{sunIrregular} & Historical Weather (T;WS;S) & Holiday, TP & - & TKDE '20 \\ 
& \emph{Ruan et al.} \cite{ruan2020dynamic} & Historical Weather (T;WS) & Holiday, TP &  POI & UbiComp '20 \\
& \emph{Luo et al.} \cite{luo2020d3p} & Historical Weather (S), AQI & Holiday, TP &  POI, Road & UbiComp '20 \\
& \emph{Liang et al.} \cite{fine_grained_2021} & Historical Weather (T;WS;S) & Holiday & POI, Road & WWW '21 \\ 

& \emph{Wang et al.} \cite{wang2021spatio} & - & Holiday &  POI & UbiComp '21 \\
& \emph{Huang et al.} \cite{huang2021multi} & Anomaly Event & - & - & UbiComp '21 \\

& \emph{Chen et al.} \cite{travel_time_cooperative_2022} & - & Holiday, TP & Road & TITS '22 \\ 
& \emph{Zhao et al.} \cite{semantic_urban_flow_2022} & Historical Weather (T;WS;S) & Holiday, TP & - & WSDM '22\\ 
& \emph{Yao et al.} \cite{MVSTGN_2023_TMC} & - & Holiday, TP & POI & TMC '23 \\ 
\hline
\end{tabular}}
\end{center}
\vspace{-1em}
\end{table*}

\begin{table}
  \small
  \caption{Data sources of contextual features. NYC for New York City; MEL for Melbourne; CA for California; `Point' and `Line' denote their geographic shapes; `N/A' indicates that the entire country generally shares the same holiday schedule.}
  \label{tab: data_source}
  \centering
  \resizebox{.95\textwidth}{!}{
  \begin{tabular}{lccccccccc}
    \toprule
    % \multirow{2}{*}{Dataset} & \multirow{2}{*}{ST Data} & Weather & Holiday & POI & \multirow{2}{*}{Application} \\
    % \cmidrule(r){3-5} 
    % & & Weather & Holiday & POI \\
    \multirow{2}{*}{\textbf{Data Source}} & \multicolumn{2}{c}{\textbf{Temporal View}} & \multicolumn{2}{c}{\textbf{Spatial View}} & \multicolumn{2}{c}{\textbf{Accessible?}} \\
    \cmidrule(lr){2-3} \cmidrule(lr){4-5} \cmidrule(lr){6-7}
     & Range & Granularity & Range & Granularity & Real-time & Historical\\
    \midrule
    \textbf{\textit{Historical Weather}} \\
    Accuweather~\cite{rahman2020short,accuweather} & undisclosed & 1 minute & Global & Avg. $5000km^2$ & \checkmark & \\
    ASOS~\cite{asos} & 1940-now & 1 hour & Global & Avg. $1000km^2$ &  & \checkmark\\ 
    NOAA~\cite{hou2021short,noaa} & 1940-now & 1 hour & America & Avg. $1000km^2$ & \checkmark & \checkmark\\ 
    Open Weather~\cite{open_weather}
    & 1979-now & 1 minute & Global & undisclosed & \checkmark & \checkmark\\
    Weather Underground~\cite{liu2019contextualized,weatherunderground}
    & 1940-now & 1 hour & Global & Avg. $5000km^2$ & \checkmark & \checkmark\\
    \midrule
    \textbf{\textit{Weather Forecast}} \\
    ECMWF~\cite{ECMWF}& 1940-now & 1 hour & Global & $28km \times 28km$& \checkmark & \checkmark \\
    GFS~\cite{gfs} & 2015-now & 3 hours & Global & $28km \times 28km$ & \checkmark & \checkmark\\
    Open Weather~\cite{open_weather}
    & 2017-now & 1 hour & Global & $2km \times 2km$ & \checkmark & \checkmark\\
    \midrule
    \textbf{\textit{AQI}} 
    \\
    Accuweather~\cite{accuweather}& undisclosed & 1 hour & Global & Avg. $500km^2$ & \checkmark & \\
    Air Now~\cite{airnow}& undisclosed & 1 hour & Global & Avg. $500km^2$ & \checkmark & \\
    EPA~\cite{epa}& 1980-now & 1 hour & USA & Avg. $500km^2$ &  & \checkmark \\
    IQAir~\cite{iqair}& undisclosed & 1 day & Global & Avg. $500km^2$ & \checkmark & \\
    WAQI~\cite{WAQI}& undisclosed & 1 hour & Global & Avg. $500km^2$ & \checkmark & \\
    \midrule
    
    \textbf{\textit{Holiday}} \\ workalendar~\cite{workalendar} &1991-now & 
 1 day & Global & N/A & \checkmark & \checkmark \\
    
    \midrule
    \textbf{\textit{POI}} \\ 
OpenStreetMap~\cite{openstreetmap} & 2010-now & 7 days & Global & Point  &  & \checkmark \\
Foursquare~\cite{foursquare}&  2009-now & 1 day & Global & Point  &  & \checkmark\\
% SateLLite-imagery-dataset & 2013-2019 & &America & City    &-  & \checkmark\\
GoogleMap~\cite{googlemap} & 2005-now & 30 days & Global & Point &  & \checkmark\\
\midrule
    
\textbf{\textit{Demographics}}\\ 
Government Website~\cite{nyc_government_website,chicago_government_website,abs_government_website}&  2010-2020 & 10 years & USA, Australia & Avg. $0.03km^2$  &  & \checkmark\\
NHGIS~\cite{NHGIS} &  1970-now & 10 years & USA & Avg. $0.03km^2$&     & \checkmark\\
\midrule
\textbf{\textit{Road Network}} \\ 
OpenStreetMap~\cite{openstreetmap} &  2010-now & 7 days & Global & Line &  & \checkmark\\
GoogleMap~\cite{googlemap} & 2005-now & 30 days & Global& Line &  & \checkmark\\
\midrule

\textbf{\textit{Administrative Division}}\\ 
Government Website~\cite{nyc_government_website,chicago_government_website} & 2000-now & hardly update & USA & Avg. $5.5km^2$ &    & \checkmark\\
    \bottomrule
  \end{tabular}}
  \vspace{-.5em}
\end{table}

\subsection{Taxonomy of Contextual Features}

\begin{figure}[htbp]
\centering
\includegraphics[width=.55\textwidth]{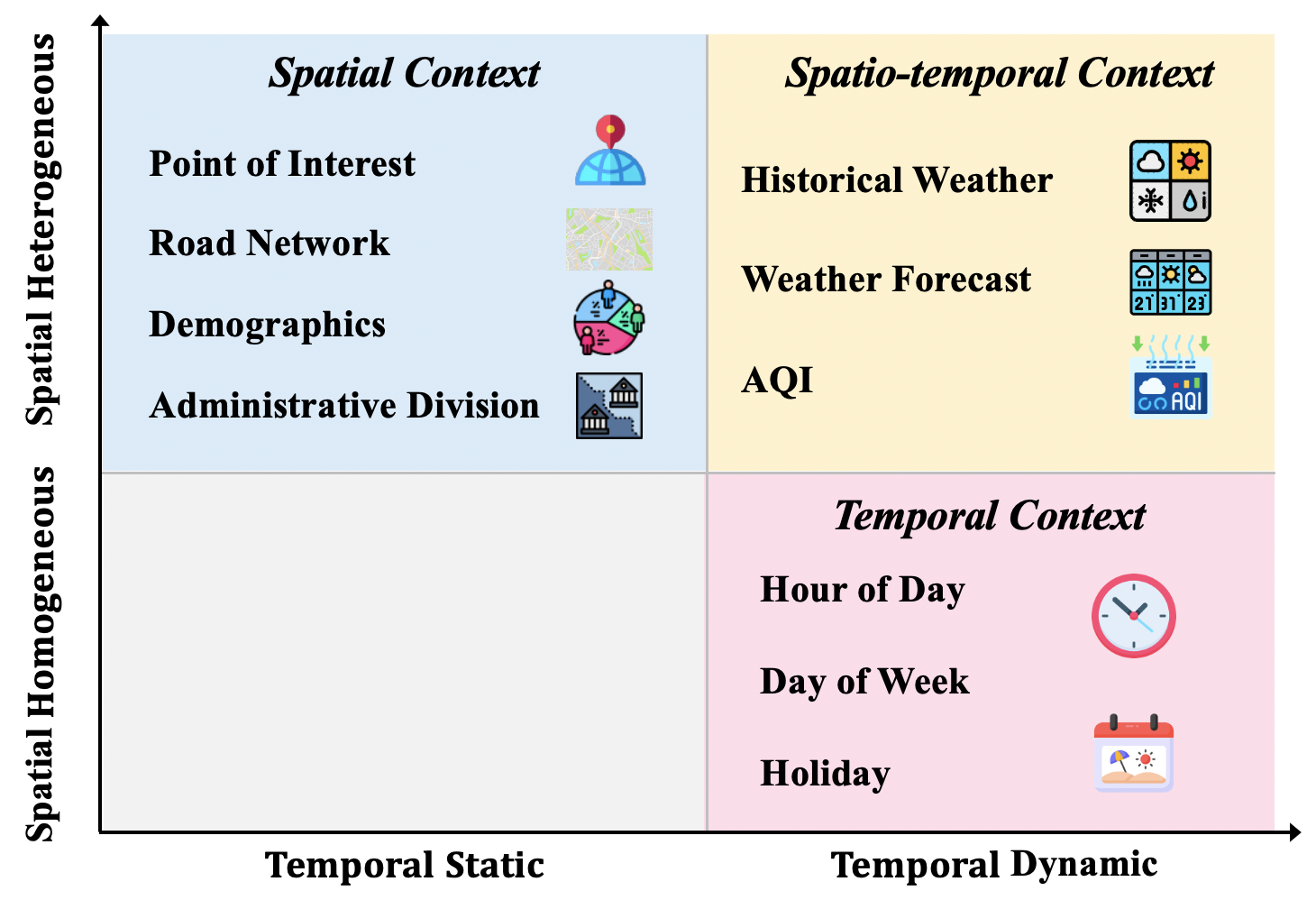}
\vspace{-1em}
\caption{Taxonomy of contextual features.}
\label{fig: taxonomy_features}
\vspace{-1em}
\end{figure}

To better understand the characteristics of different types of context, we propose an extensible context taxonomy. The primary challenge in crowd flow prediction stems from complex temporal dynamics and spatial heterogeneity~\cite{STDM_problem_2018}, where context can provide extra information for temporal or spatial modeling.
Based on this insight, we highlight the spatio-temporal properties of the context and classify contextual features into three categories: temporal context, spatial context, and spatio-temporal context, as shown in Figure~\ref{fig: taxonomy_features}.
Temporal contextual features, such as holidays, are dynamic in time and homogeneous in space, which implies that they remain consistent across various locations during the same period. We refer to these features as $\mathcal{T}=\{\tau_1, \tau_2,...\}$. Each element in $\mathcal{T}$ includes a timestamp and observation value (e.g., 2014-11-27, Thanksgiving Day). Spatial contextual features are static in time and heterogeneous in space, while spatio-temporal contextual features vary over both time and space. These are represented as $\mathcal{S}=\{s_1, s_2, \ldots\}$ and $\mathcal{ST}=\{e_1, e_2, \ldots\}$, respectively. Each $\mathcal{S}$ element includes a location and observation value, while each $\mathcal{ST}$ element includes a timestamp, location, and observation value.

\textbf{Weather and AQI} are categorized as spatio-temporal contextual features in our taxonomy. Historical weather and AQI data are typically collected from meteorological stations equipped with multiple sensors~\cite{djordjevic2019smart,colston2018evaluating}. For instance, in New York City, there are three weather monitoring stations and over ten pollutant monitoring stations. Taking the weather station named LGA as an example, the LGA station reports 20+ variables at around the 51st minute every hour. Forecast weather data is generated by numerical prediction models that operate at specific intervals with a fixed spatio-temporal granularity. For instance, the Global Forecast System (GFS) runs its prediction model four times daily at 00:00, 06:00, 12:00, and 18:00 UTC, producing forecasts at 3-hour intervals on predefined grids (e.g., $28 km \times 28 km$).

\textbf{Holiday, Temporal Position} are categorized as temporal contextual features in our taxonomy, as the regions of interest generally share the same time zone and holiday schedule. The Holiday feature indicates whether a day is a holiday, while the temporal position feature typically includes \textit{HourofDay} and \textit{DayofWeek}~\cite{context_generalizability}, distinguishing different hours (e.g., 8 am vs. 9 pm) and days (e.g., Monday vs. Saturday).

\textbf{POI, Road Network, Demographics, Administrative Division,} and \textbf{Spatial Position} are categorized as spatial contextual features in our taxonomy. These features are updated less frequently than weather or holiday data. For instance, in New York City, the number of POIs changed by about 5\% between 2014 and 2015, while the road count changed by 15\%. Spatial position features distinguish different spatial units~\cite{wang_deepsd_2017,han_traffic_activity_2023} and are generally considered static over time. Demographic and administrative division data are updated only every few years.

\subsection{Data Source Investigation}
To comprehensively collect the reliable context data selected in Section~\ref{sec: context_selection}, we investigate various data sources through previous research and online resources. Table~\ref{tab: data_source} summarizes the sources of different contextual features, their spatio-temporal ranges and granularity, and their accessibility (whether real-time or historical data is available). Note that we define the spatial granularity of historical weather and AQI based on the entire city area relative to the number of meteorological stations, ensuring it approximately represents the monitoring range. For instance, as shown in Table~\ref{tab: data_source}, an ASOS meteorological station typically monitors weather states over 1000 $km^2$. For weather forecasts, we record the size of the basic grid used in the numerical weather forecasting model \cite{gfs, ECMWF}. We also report the average area covered by spatial contextual features, represented as geographic polygons (e.g., administrative divisions or census tracts for demographic statistics).

When collecting context from these resources, we filtered out some data sources that did not meet specific criteria. For historical weather and forecasts, we excluded Accuweather and Open Weather due to their lack of historical archives (i.e., only providing real-time data) and the additional fees required, respectively. We also chose not to use GFS because of its shorter temporal range (i.e., accessible from 2015 onward) and coarser temporal granularity (i.e., 3 hours) compared to ECMWF. For AQI, we selected EPA as it is the only source with accessible historical records. Additionally, we excluded POI and Road Network sources that required extra fees, such as Foursquare and GoogleMap. No exclusions were made for demographics and administrative division sources. 

As we exclude certain resources for building \texttt{STContext}, we encourage users to carefully consider task requirements and select appropriate data sources. For instance, if accurate and comprehensive commercial POI data is needed, GoogleMap may be a better choice than OpenStreetMap. We believe the table of investigated data sources will benefit a wide range of researchers.

\subsection{Multi-Source Data Fusion Procedure}
After investigating data from various sources, we found multiple qualified options for historical weather (e.g., ASOS and NOAA). Unlike most previous work that relied on a single source \cite{zhang2017deep, yuan_demand_2021}, we argue that collecting and fusing data from multiple sources can enhance dataset quality and propose a multi-source data fusion procedure. 

\begin{figure}[htbp]
\centering
\includegraphics[width=.98\textwidth]{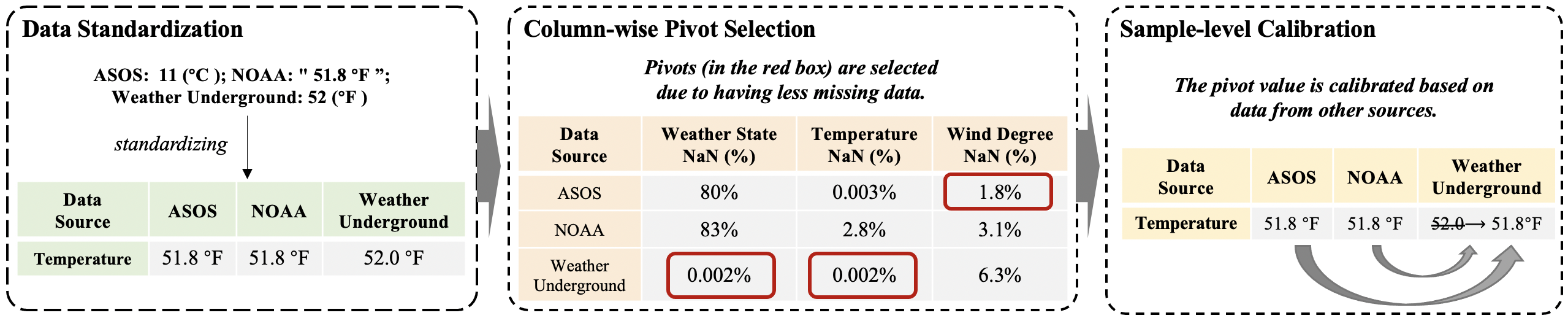}
\vspace{-.5em}
\caption{Illustrative example of the proposed multi-source data fusion procedure.}
\label{fig: multi_source_fusion}
% \vspace{-.5em}
\end{figure}

Figure \ref{fig: multi_source_fusion} illustrates our multi-source data fusion procedure, which consists of three steps. The first step, data standardization, unifies data formats from different sources for better comparability. The second step is column-wise pivot selection, which leverages the advantages of various data sources. This step is motivated by the strengths of data sources in specific feature fields. For instance, the ASOS dataset from NYC (from July 1, 2013, to September 28, 2017) shows an over 80\% missing rate for weather states but under 2\% for wind degrees. We assume that data sources with less missing data offer higher quality, so we select pivots with minimal missing data for each feature field. These pivot values serve as defaults for the fused data.
The third step is sample-level calibration, which adjusts the pivot value for each sample when it may be inaccurate. Two situations may render the pivot value inappropriate: first, when the pivot value is NaN, and second, when it conflicts with values from other sources (e.g., as shown in the right chart of Figure \ref{fig: multi_source_fusion}, ASOS and NOAA share the same temperature value, while Weather Underground, the pivot source, records a different value). By incorporating data from other sources (e.g., by averaging or majority voting), the quality of the pivot value can be improved.

\section{\texttt{STContext} Dataset} 

\subsection{Basic Statistics}

As shown in Table~\ref{table: dataset_statistical}, the \texttt{STContext} comprises 9 crowd flow datasets spanning 4 months (e.g., Vehicle Speed dataset in BAY) to 7 years (e.g., Ride-sharing in NYC) across 5 tasks and 6 cities, ensuring a rich diversity in both city coverage and time span. We collected 10 contextual features matched to each crowd flow dataset's spatio-temporal range, except for the pedestrian dataset in Melbourne (due to data source limitations shown in Table~\ref{tab: data_source}). Table~\ref{table: dataset_statistical} provides details on the crowd flow datasets and their associated contextual features (excluding spatial and temporal positions that can be generated). For researchers’ convenience, we summarize the dimensionality, missing rate, and number of spatial units for spatio-temporal context, the number of events for temporal context, as well as the number of classes and records for spatial context.

\begin{table}[htbp]
\small
    \caption{\texttt{STContext} dataset statistics. Wea. stands for Weather. \#dim stands for the number of contextual feature dimensions, \#station represents the number of monitoring stations, and NARate stands for the average missing rate of each column of contextual features. \#class stands for the number of possible values of the contextual features, while \#record represents the total number of data records. \# Holiday stands for the number of holidays. We did not find available AQI data for Melbourne. NYC for New York City; MEL for Melbourne; BAY for San Francisco Bay Area.}
    \resizebox{0.99\textwidth}{!}{
    \setlength{\tabcolsep}{.5mm}{
    
    \begin{tabular}{cccccccccccccccccccc}
\toprule
\multirow{2}{*}{\textbf{Task}} & \multirow{2}{*}{\textbf{City}} & \multirow{2}{*}{\textbf{Time Span}} & \multicolumn{2}{c}{\textbf{Historical Wea.}} & \multicolumn{2}{c}{\textbf{Wea. Forecast}} & \multicolumn{2}{c}{\textbf{AQI}} & \multirow{2}{*}{\textbf{\# Holiday}} & \multicolumn{2}{c}{\textbf{POI}} & \multicolumn{2}{c}{\textbf{Road}} & \multicolumn{2}{c}{\textbf{Demographics}} & \multicolumn{1}{c}{\textbf{A.D.}} \\
\cmidrule(lr){4-5} \cmidrule(lr){6-7} \cmidrule(lr){8-9} \cmidrule(lr){11-12} \cmidrule(lr){13-14} \cmidrule(lr){15-16} \cmidrule(lr){17-17}
& & & \#dim & NARate & \#dim &\#units & \#dim & \#units & & \#class&\#record &\#class&\#record&\#class&\#record&\#record\\

\midrule
\multirow{3}{*}{Bike-sharing} & NYC & 2013/07-2017/09 & 25 & 0.7447 & 12 & 117 & 5 & 43 & 493 & 35 &16936&  27 &3499 & 56 & 37991 & 262\\
~ & Chicago & 2013/07-2017/09 & 25 & 0.7435 & 12 & 117 & 5 & 48 & 486 & 35 & 6137&27 &2704 & 56 & 46292 & 98\\
~ & DC & 2013/07-2017/09 & 25 & 0.7372 & 12 & 81 & 5 & 4 & 490 & 35 &13434&27 &73050 & 56 & 6014 & 46\\

\midrule
Pedestrian & Melbourne & 2021/01-2022/11 & 25 & 0.6811 & 12 & 81 & N/A & N/A & 212 & 35 &26712 &27 &6187 & 2 &11034 & 17\\

\midrule
\multirow{2}{*}{Vehicle Speed} & LA & 2012/03-2012/06 & 25 & 0.6966 & 12 & 117 & 5 & 205 & 36 & 35 &1731& 27 & 8205 & 56 &253596 & 183\\
~ & BAY & 2017/01-2017/07 & 25 & 0.7422 & 12 & 221 & 5 & 204 & 86 & 35 &3209 &27 &42788 & 56 &77921 & 23\\

\midrule
\multirow{2}{*}{Ride-sharing} & Chicago & 2013/01-2022/03 & 25 & 0.7435 & 12 & 117 & 5 & 48 & 1081 & 35 &11636& 27 &2324 & 56 & 65021 & 46\\
~ & NYC & 2016/01-2023/06 & 25 & 0.7447 & 12 & 117 & 5 & 42 & 871 & 35 &49255& 27 &29140 & 56 &37991 &262\\

\midrule
Metro & NYC & 2022/02-2023/12 & 25 & 0.7423 & 12 & 117 & 5 & 39 & 212 & 35 &49255& 27 &29140 &56 &37991&262  \\
\bottomrule
    \end{tabular}}}
    \label{table: dataset_statistical}
\end{table}

\subsection{Data Analysis}

\begin{figure}[htbp]
\begin{minipage}[c]{0.48\linewidth}
    \centering
    \includegraphics[width=1\textwidth]{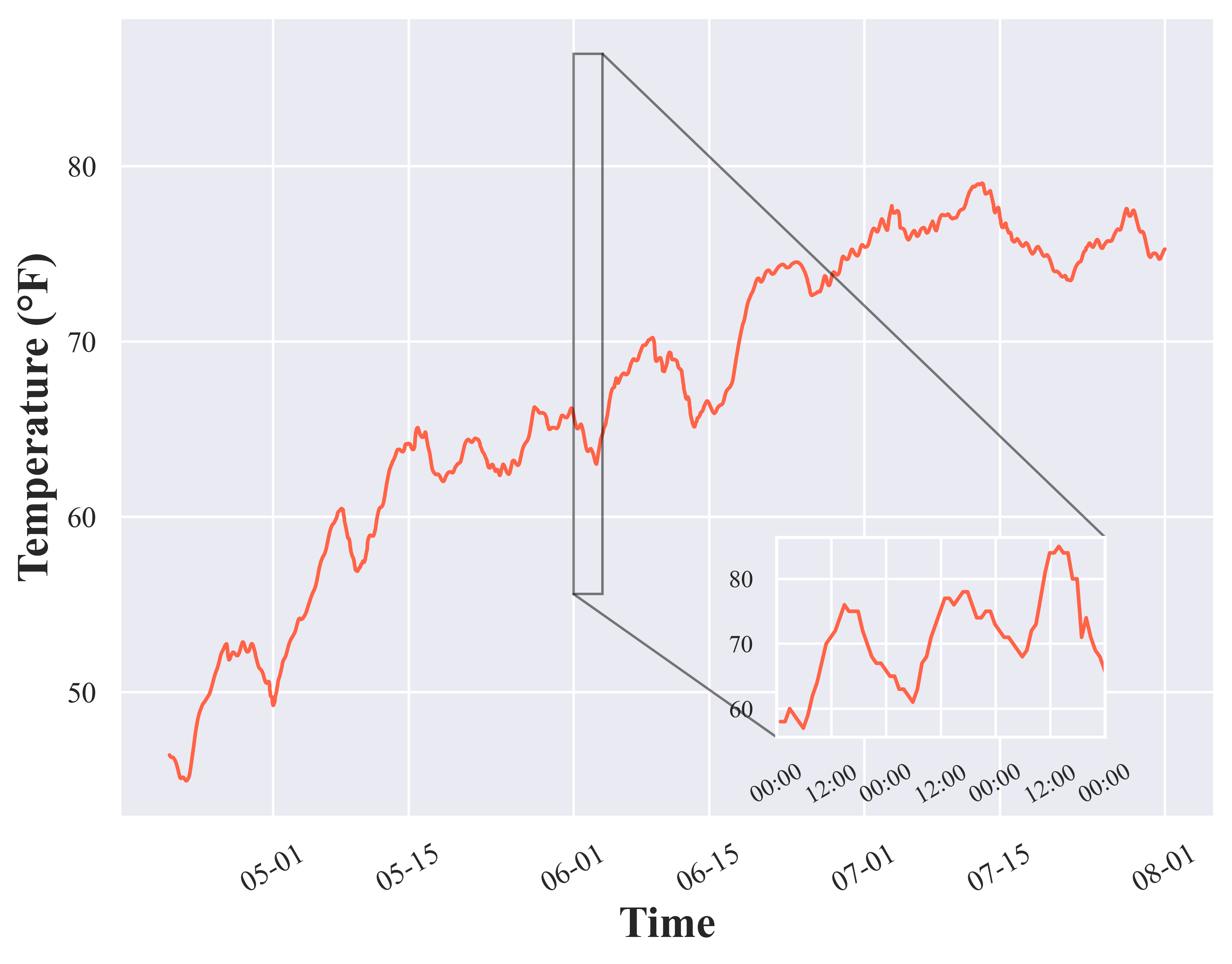}
    \vspace{-2em}
    \caption{Long-term trend and short-term periodicity.}
    \label{fig: temporal_dynamics}
\end{minipage}
\hspace{.8em}
\begin{minipage}[c]{0.48\linewidth}
\centering
\includegraphics[width=1\textwidth]{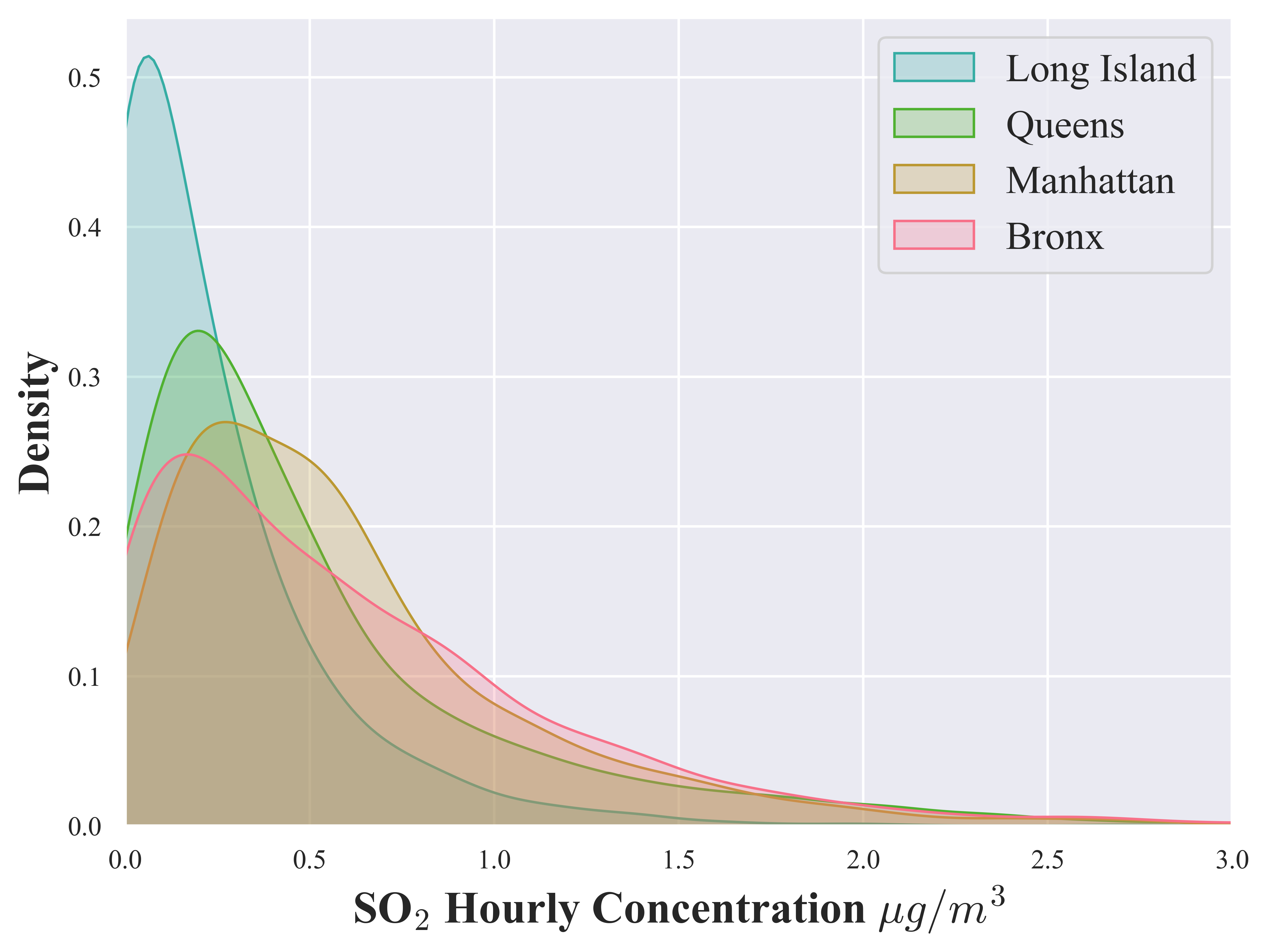}
\vspace{-1.2em}
\caption{Spatial heterogeneity of SO$_2$}
\label{fig: spatial_hetero}
\end{minipage}
\end{figure}

\subsubsection{Variations in time and space} 
Modeling context features is complex due to inherent temporal dynamics and spatial heterogeneity, as discussed below:

\textbf{Temporal Dynamics}. The temporal dynamics of contextual features are complex, often displaying long-term trends and short-term periodicity due to natural laws~\cite{holton2013introduction}. For example, as shown in Figure~\ref{fig: temporal_dynamics} in New York City temperature data from May to August 2014, the temperature rises gradually with seasonal change and fluctuates daily due to day-night cycles. These patterns result from the Earth’s revolution, driving seasonal shifts, and its rotation, producing daily cycles.

\begin{figure}[htbp]
\begin{minipage}[c]{0.48\linewidth}
    \centering
    \includegraphics[width=1\textwidth]{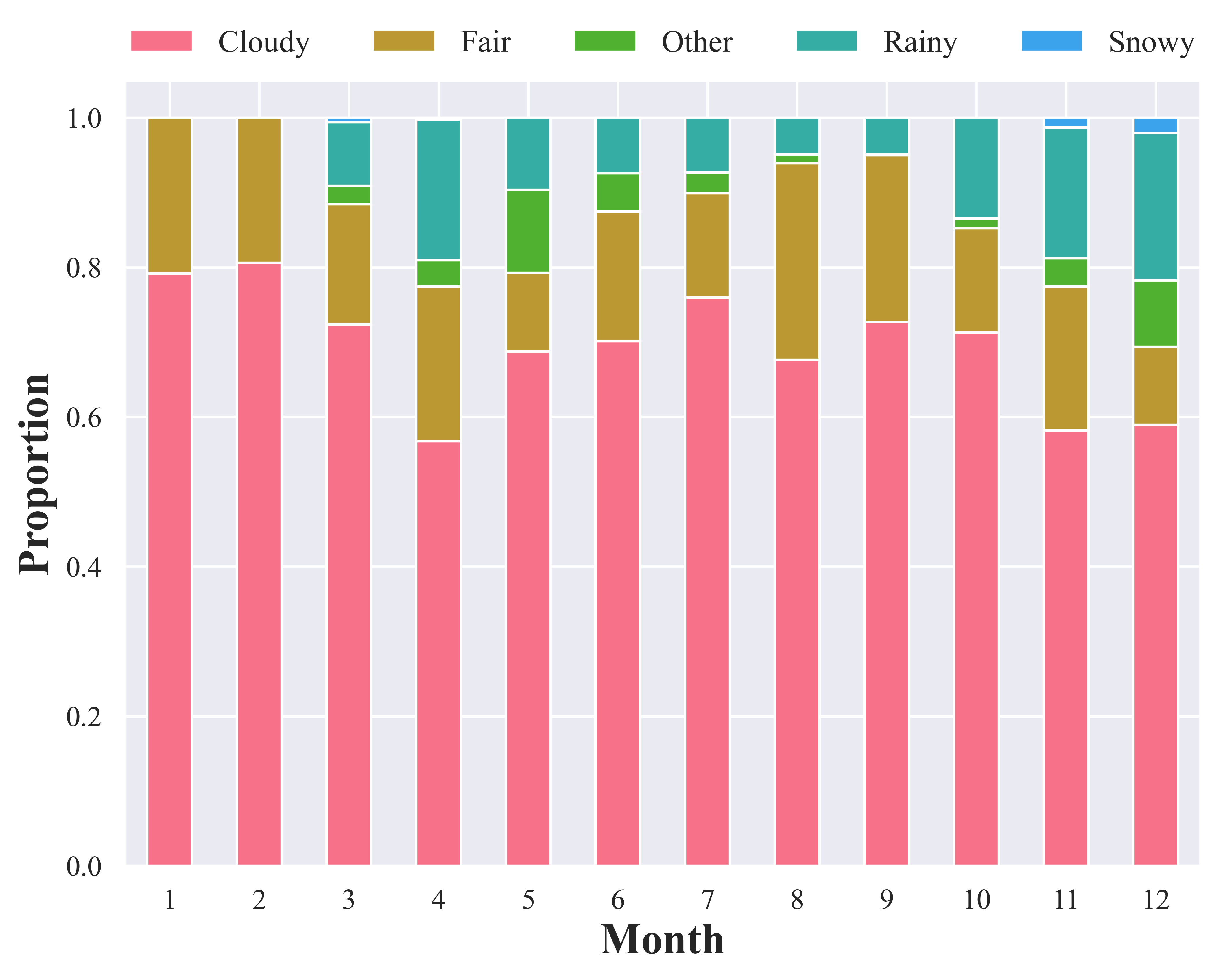}
    \vspace{-2.5em}
    \caption{Weather distribution over time dimension.}
    \label{fig: temporal_distribution}
\end{minipage}
\hspace{0.9em}
\vspace{-1em}
 \begin{minipage}[c]{0.48\linewidth}
    \centering
    \includegraphics[width=0.95\textwidth]{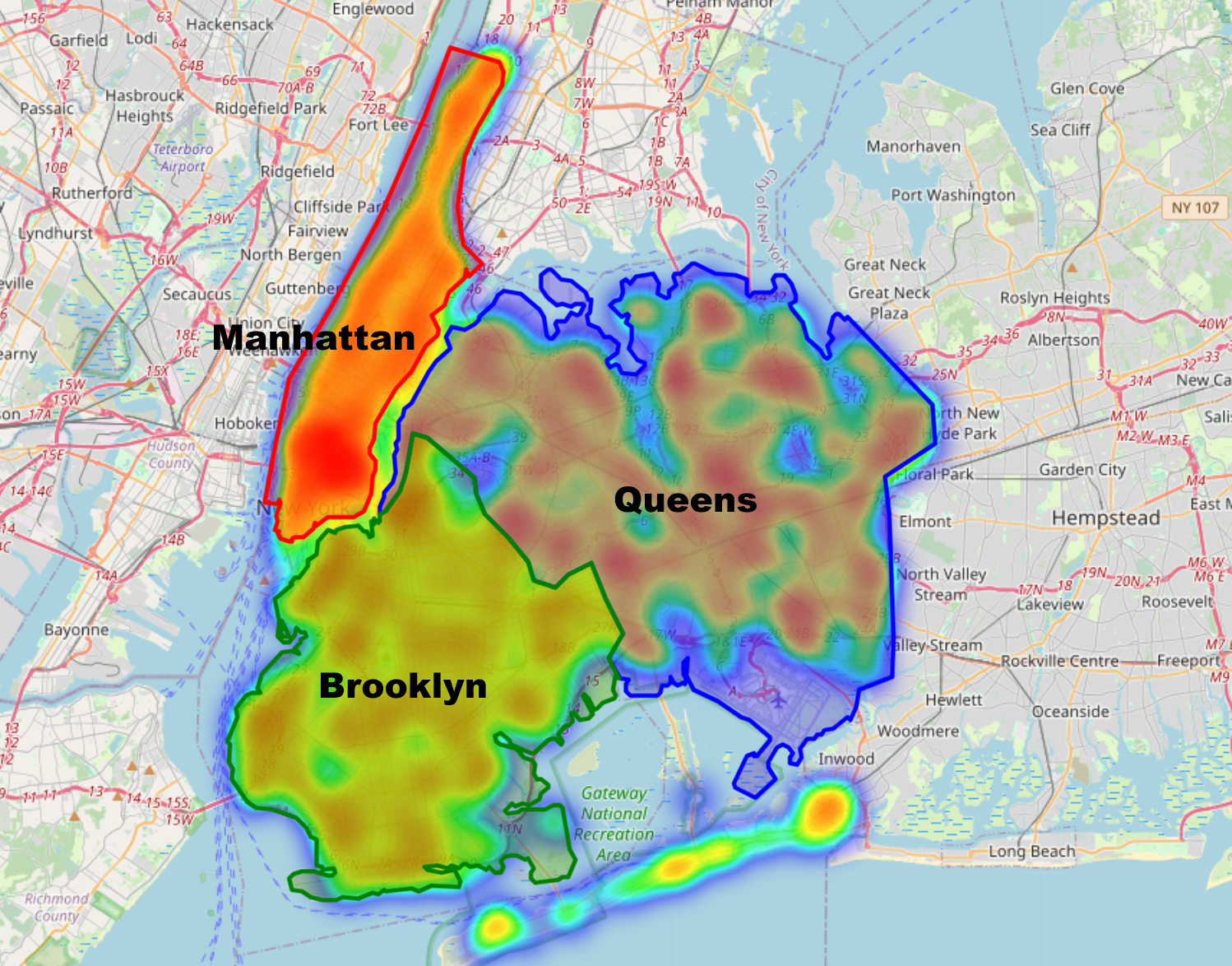}
    \vspace{-.5em}
    \caption{POI distribution over space dimension.}
    \label{fig: spatial_distribution}
\end{minipage}   
\end{figure}

\textbf{Spatial Heterogeneity}. At the city scale, contextual features demonstrate spatial heterogeneity. As illustrated in Figure~\ref{fig: spatial_hetero}, data from 43 AQI monitoring stations in New York City over one year shows that the average relative concentration of SO$_2$ is lowest in Long Island, followed by Queens, while Manhattan and the Bronx exhibit higher SO$_2$ concentrations. This spatial heterogeneity may stem from the more developed economy and the greater number of factories in the city center, contributing to higher SO$_2$ emissions compared to the suburbs.

\subsubsection{Imbalanced distribution of contextual features}
Contextual features are often unevenly distributed across time and space. For instance, as shown in Figure~\ref{fig: temporal_distribution}, snowfall in NYC primarily occurs from November to December, while rain occurs mainly from March to January and October to November, both representing a relatively low proportion of the total period. Similarly, as illustrated in Figure~\ref{fig: spatial_distribution}, NYC’s POI density is higher in central areas (e.g., Manhattan) and lower in surrounding suburbs (e.g., Brooklyn, Queens). These imbalanced distributions above suggest a need for more fine-grained evaluation scenarios from both temporal and spatial dimensions beyond overall metrics, as later discussed in empirical studies (see Section~\ref{sec: evaluation_on_specific_metrics}).

\begin{figure*}[htbp]
\begin{minipage}[c]{0.46\linewidth}
    \centering
    \includegraphics[width=1\textwidth]{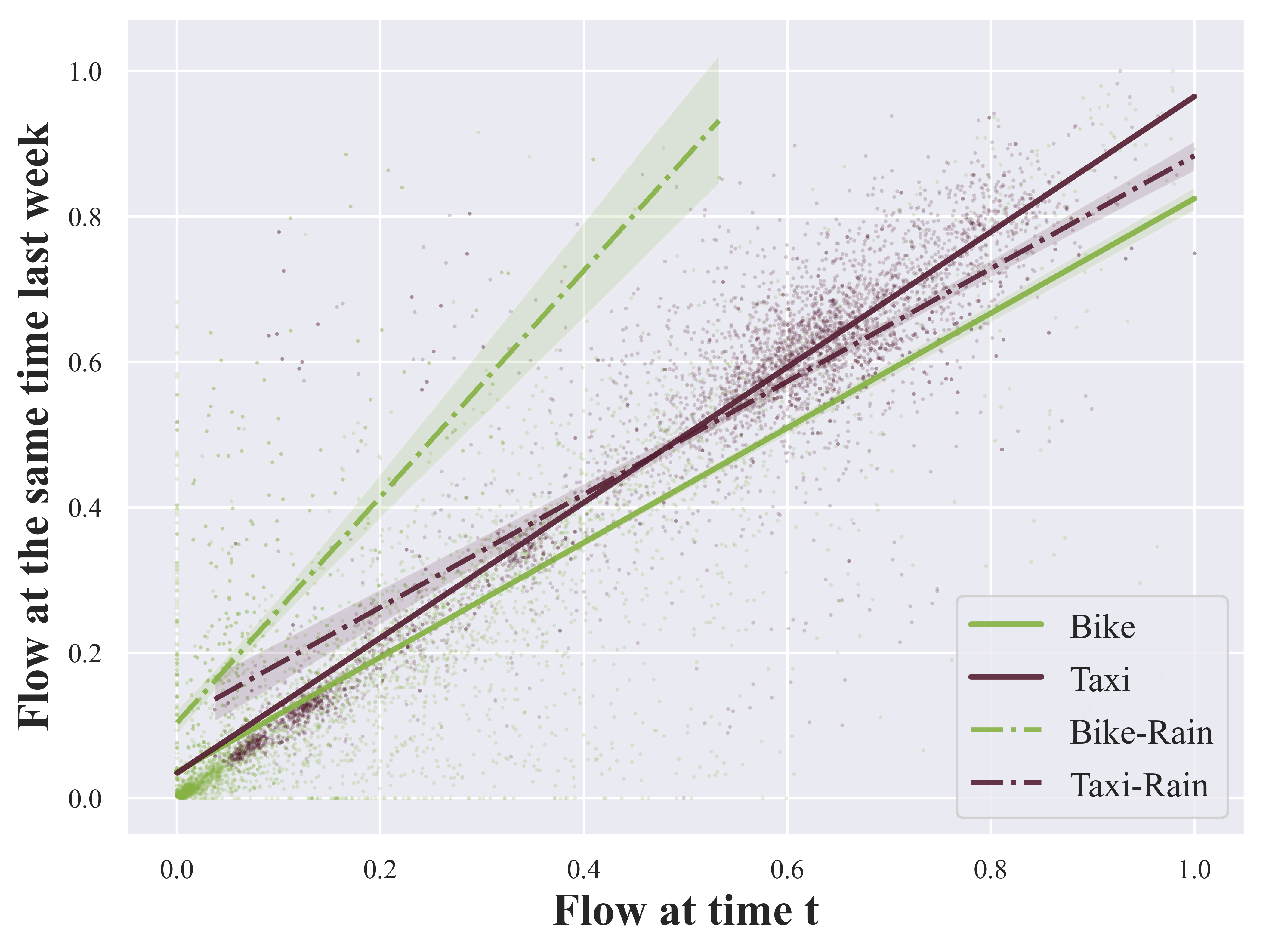}
    \vspace{-2em}
    \caption{Impact of rain states on bike and taxi flows.}
    \label{fig: rain_bike_taxi_correlation}
\end{minipage}
\hspace{.8em}
\vspace{-1em}
\begin{minipage}[c]{0.46\linewidth}
\centering
\includegraphics[width=1\textwidth]{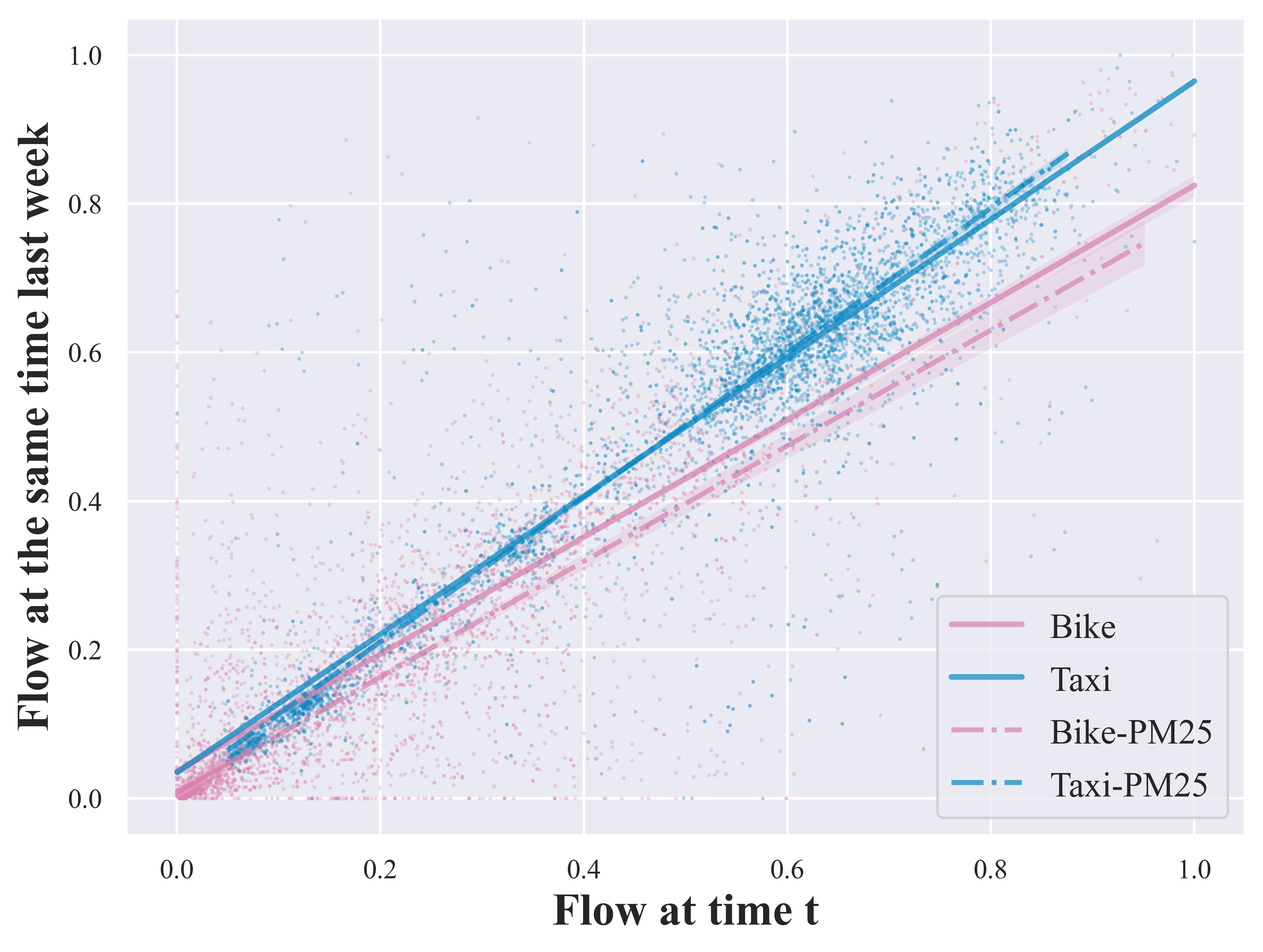}
\vspace{-2em}
\caption{Impact of air pollution on bike and taxi flows.}
\label{fig: pm25_correlation}
\end{minipage}   

\vspace{-2em}
\end{figure*}
\begin{figure*}[htbp]
\begin{minipage}[c]{0.46\linewidth}
\centering
\includegraphics[width=1\textwidth]{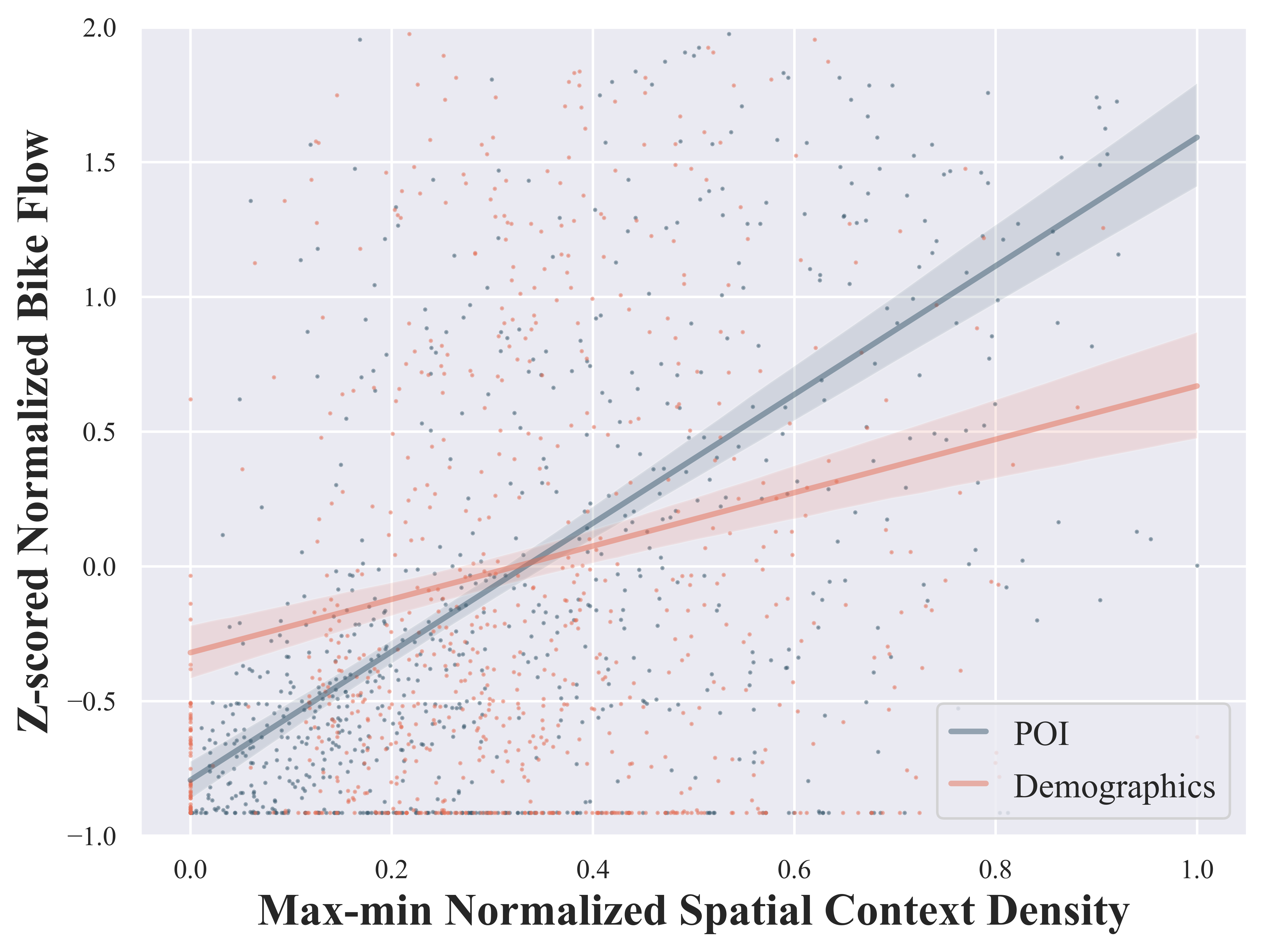}
\vspace{-2em}
\caption{Correlations between spatial contextual features (i.e., POI and demographics) and bike flow.}
\label{fig: corr_flow_poi_demo}
\end{minipage}   
\hspace{.8em}
\begin{minipage}[c]{0.46\linewidth}
\centering
\includegraphics[width=1\textwidth]{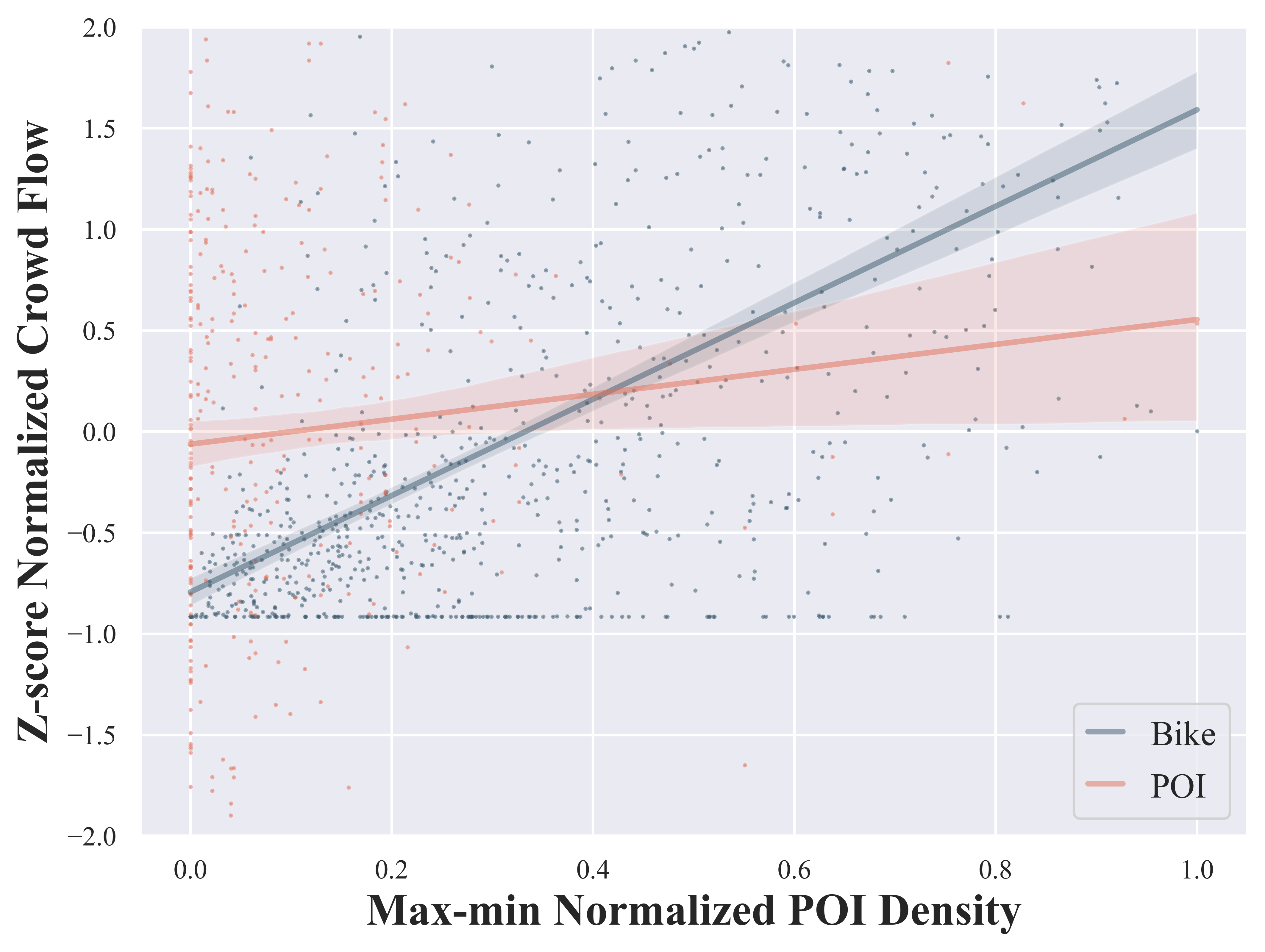}
\vspace{-2em}
\caption{Correlations between POI and crowd flow (i.e., bike flow and traffic speed flow).}
\label{fig: corr_flow_bike_speed}
\end{minipage}
% \vspace{-1em}
\end{figure*}

\subsubsection{Correlations between contextual features and crowd flow}\label{sec: context_vs_crowdflow}

The influence of contextual features on crowd flow varies by both context type and flow type.
Figure~\ref{fig: rain_bike_taxi_correlation} and Figure~\ref{fig: pm25_correlation} illustrate crowd flows during specific weather conditions or severe air pollution events (PM 2.5 above $40\mu g/m^3$ per hour~\cite{degaetano2004temporal}) compared to flows from previous weeks, showing a clear correlation due to high weekly periodicity \cite{zhang2017deep, STMeta}. When weather or air quality affects mobility, the flow curve steepens or flattens.
These figures reveal that rain significantly impacts bike flow but has less effect on taxi flow. In contrast, air pollution has minimal influence on both bike and taxi crowd flows.
Moreover, Figure~\ref{fig: corr_flow_poi_demo} shows the correlation between spatial contextual features (POI and demographics) and bike flow. 
We find a moderate positive correlation between POI and bike flow (Pearson coefficient: 0.55), while demographics show a weak correlation (Pearson coefficient: 0.18). Figure~\ref{fig: corr_flow_bike_speed} shows the relationship between POI and crowd flow (bike and traffic speed flow). POI exhibits a moderate positive correlation with bike flow (Pearson coefficient: 0.55) but shows no significant association with traffic speed flow (Pearson coefficient: 0.09).

\section{A Unified Paradigm for Contextual Feature Incorporation}
\subsection{Overview}

\begin{figure*}[htbp]
\begin{minipage}[c]{0.55\linewidth}
    \centering
    \includegraphics[width=1\textwidth]{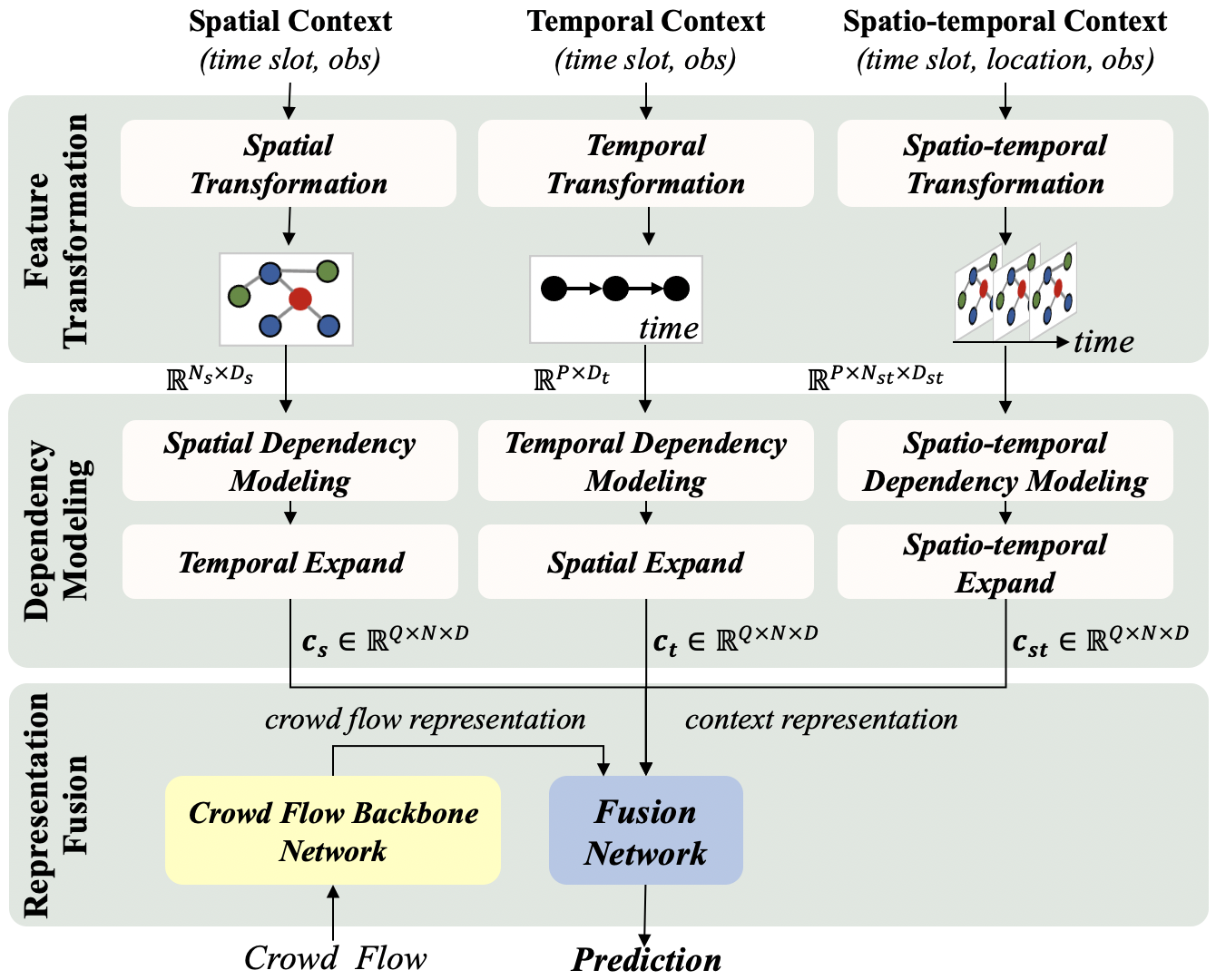}
    % \vspace{-2em}
    \caption{A unified paradigm for incorporating contextual features into STCFP methods.}
    \label{fig: modeling_paradigm}
\end{minipage}
\hspace{.5em}
 \begin{minipage}[c]{0.36\linewidth}
    \centering
    \includegraphics[width=1\textwidth]{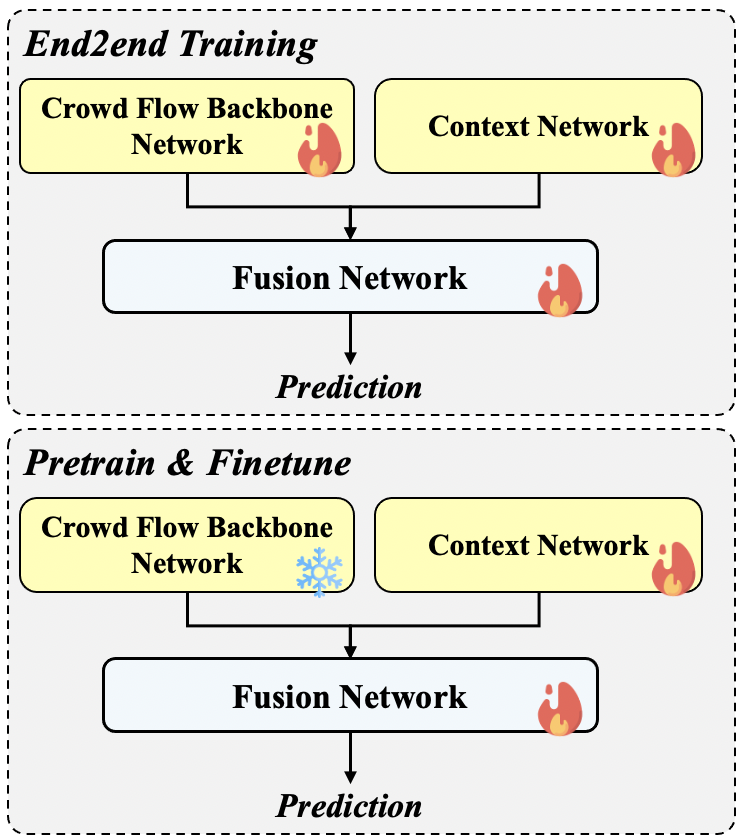}
    \vspace{-1em}
    \caption{Illustration of two training strategies.}
    \label{fig: training_strategies}
\end{minipage}   
% \vspace{-1em}
\end{figure*}

We decompose the STCFP context incorporation process into four components (Figure~\ref{fig: modeling_paradigm} and Figure~\ref{fig: training_strategies}): (a) feature transformation, (b) dependency modeling, (c) representation fusion, and (d) training strategies. 
First, \textbf{feature transformation} associates the raw contextual data with the predicted crowd flow locations, such as assigning weather data from the nearest meteorological station to each crowd flow location. 
We denote the transformed spatial, temporal, and spatio-temporal contextual features as $\mathbf{SC}\in \mathbb{R}^{N_s\times D_s}$, $\mathbf{TC}\in \mathbb{R}^{P\times D_t}$, and $\mathbf{STC}\in \mathbb{R}^{P\times N_{st}\times D_{st}}$ ($D_{s}$, $D_{t}$, $D_{st}$ are the feature dimensions). 
Second, \textbf{dependency modeling} employs neural networks (e.g., RNNs for temporal dependency modeling \cite{ke_short-term_2017}) to learn context representations, which are then expanded along the temporal or spatial axes to align with the crowd flow feature map.
Third, the expanded context and crowd flow representations are fused to make more comprehensive and informative representations, namely \textbf{representation fusion}.
Finally, various \textbf{training strategies} (e.g., end-to-end training) can be applied to learn the traffic and context representations either together or independently.

\begin{table}
  \small
  \caption{Summary of representative STCFP models in the proposed paradigm. `TP' refers to the temporal position feature (e.g., time of day). The symbol `$\times$' indicates the combination of spatial and temporal contextual features, which together form a spatio-temporal context, as they exhibit both temporal dynamics and spatial heterogeneity. }
  \label{tab: representative_models}
  \centering
  \resizebox{.95\textwidth}{!}{
  \begin{tabular}{lcccccccccc}
\toprule
\multirow{2}{*}{\textbf{Method}} & \multirow{2}{*}{\textbf{Context Type}} & \textbf{Transformation} & \textbf{Modeling} & \textbf{Fusion} & \textbf{Training} \\

& ~ & \textbf{Hypothesis} & \textbf{Hypothesis} & \textbf{Techniques} & \textbf{Strategies} \\

\midrule
\multirow{2}{*}{\textit{ST-ResNet}~\cite{zhang2017deep}
} & Weather (ST) & Proximity and Closeness & \textit{Known and Space-invariant} & \multirow{2}{*}{\textit{Add}} &  \multirow{2}{*}{\textit{End2End}}  \\
% \cmidrule(lr){2-4}
& Holiday (T) & Closeness & \textit{Only Known} & & \\
\midrule
\textit{Multi-graph~\cite{chai_multi_graph_2018}
} & Weather\&Holiday (T) & Closeness & \textit{Known} & \textit{Concat} & \textit{Pretrain \& Finetune} \\
\midrule
\textit{DeepSTD}~\cite{zheng_deepstd_2020} & Weather\&Holiday $\times$ POIs (ST) & Proximity and Closeness & \textit{Known and Space-varying} & \textit{Add} & \textit{End2End} \\
\midrule
\textit{MVGCN}~\cite{sunIrregular} & Weather\&Holiday\&TP (T) & Closeness & \textit{Only Known} & \textit{Gating} & \textit{End2End}\\
\midrule
\textit{STRN}~\cite{STRN_2021} & Weather\&Holiday $\times$ POIs (ST) & Proximity and Closeness & \textit{Known and Space-invariant} & \textit{Concat} & \textit{End2End} \\
\midrule
\textit{ST-GSP}~\cite{st_gsp_2022}  & Weather\&Holiday\&TP (T) & Closeness & \textit{Only Known} & \textit{Add} & \textit{End2End} \\
\midrule
\textit{MVSTGN}~\cite{MVSTGN_2023_TMC} & Holiday\&TP $\times$ POIs (ST) & Proximity and Closeness & \textit{Known and Space-invariant} & \textit{Add} & \textit{End2End} & \\
% ~ & POIs (S) & Proximity  &  Space-invariant & ~ & ~ & \\
\bottomrule
  \end{tabular}}
  % \vspace{-1em}
\end{table}

Note that we do not include model-specific design details, such as the number of stacked RNN layers for temporal modeling. As far as we know, the four dimensions encompass a broad range of options representative of the open literature. To confirm this, Table \ref{tab: representative_models} provides a summary of representative
methods, discussing how they align with our proposed paradigm.

\subsection{Key Components and Design Choices} 
We discuss the basic assumptions of four key components, including feature transformation, dependency modeling, fusion techniques, and training strategies, along with several representative options for each.

\subsubsection{Feature Transformation} 
The purpose of feature transformation is to retrieve useful contextual features for predicting crowd mobility, typically as part of data pre-processing \cite{deep_fusion_net_2018, context_generalizability}.  
Most existing studies, as discussed later, adopt a similar transformation hypothesis (i.e., closeness or proximity). While they are reasonable, we believe that feature transformation has been somewhat overlooked and warrants more careful consideration.
We here elaborate on three types of transformations and their underlying hypotheses.

\begin{figure}[h]
\begin{minipage}[c]{0.43\linewidth}
    \centering
    \includegraphics[width=1\textwidth]{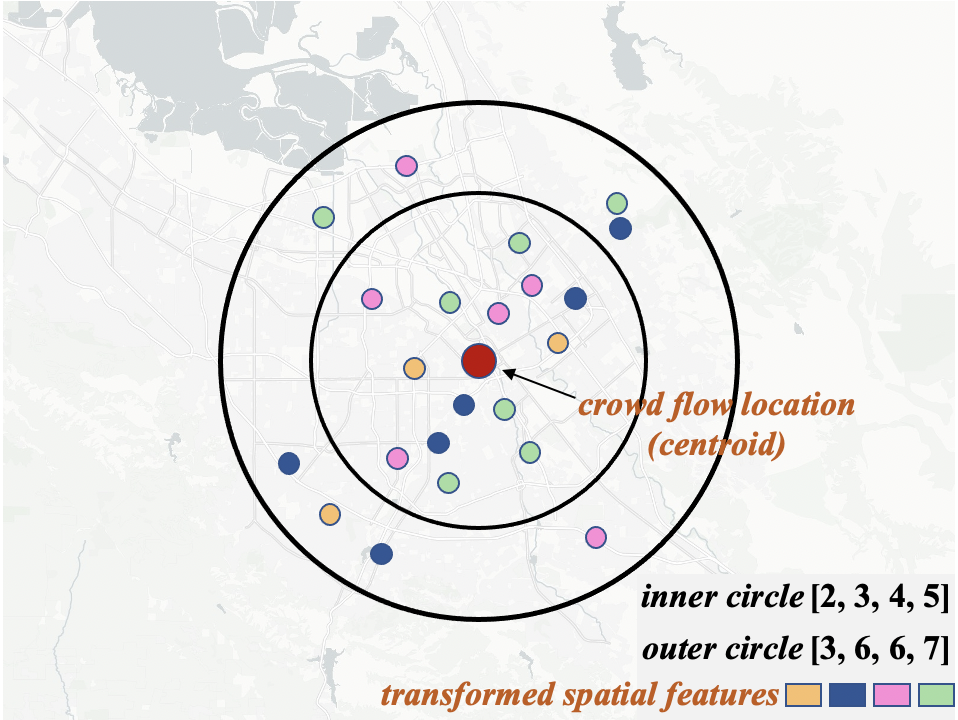}
    \vspace{-1.35em}
    \caption{Illustration of spatial transformation.}
    \label{fig: spatial_transformation}
\end{minipage}
\hspace{.5em}
 \begin{minipage}[c]{0.45\linewidth}
    \centering
    \includegraphics[width=1\textwidth]{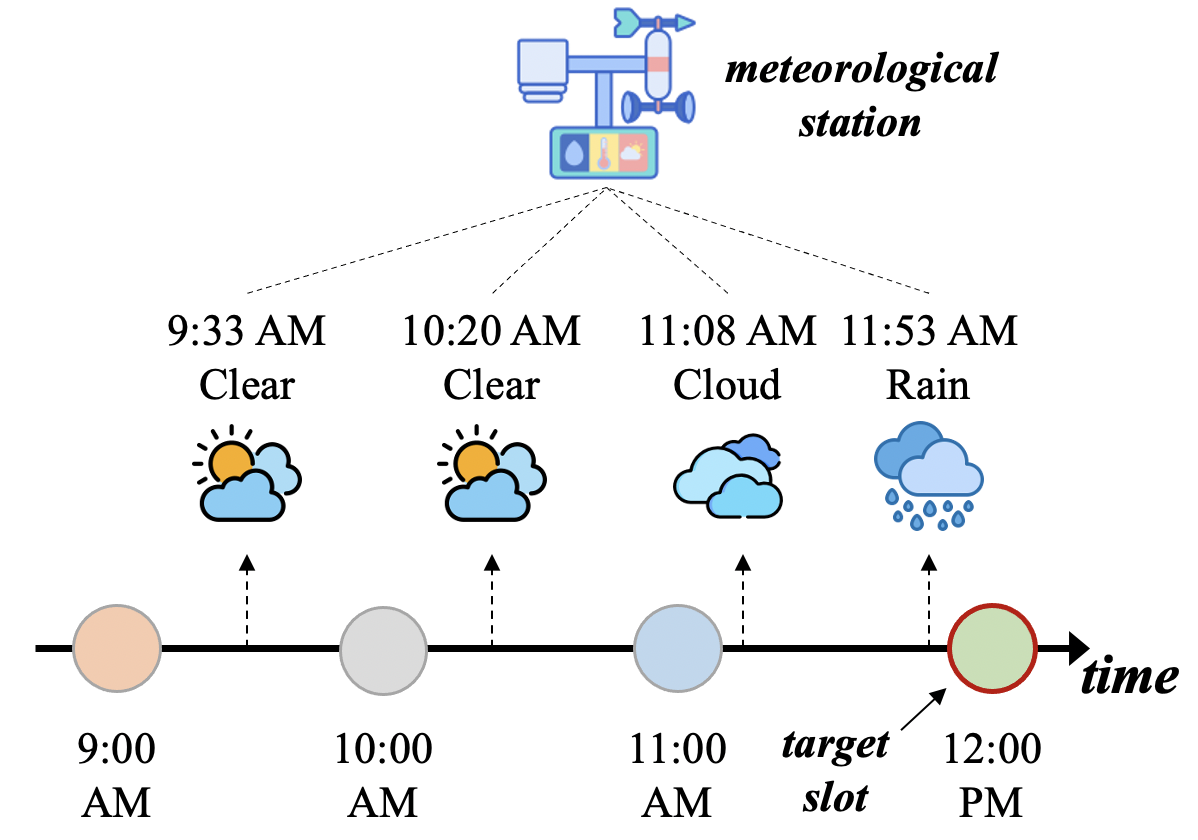}
    % \vspace{-1em}
    \caption{Illustration of temporal transformation.}
    \label{fig: temporal_transformation}
\end{minipage}   
% \vspace{-1em}
\end{figure}

\begin{itemize}[leftmargin=2em]
    \item \textit{Spatial transformation:} Most methods focus on the idea that things closer together are more related (i.e., proximity), as suggested by Waldo R. Tobler in 1969: `Everything is related to everything else, but near things are more related than distant things.' Accordingly, nearby context can be treated as spatial properties of the predicted crowd flow stations \cite{deep_fusion_net_2018,airformer_2023}. In practice, we first identify a spatial area where the contexts are assumed to influence the predicted crowd flow. Then, we convert the context within this area into spatial features using techniques like summation \cite{chai_multi_graph_2018} or distance-based decay \cite{deep_fusion_net_2018,airformer_2023}. Figure~\ref{fig: spatial_transformation} shows an example of spatial transformation, where the inner and outer circles represent different regions affecting crowd mobility. Finally, we aggregate four types of contextual features by counting.
    
    \item \textit{Temporal transformation:} Contextual features from nearby times exhibit temporal autocorrelation \cite{zhang2017deep,STDM_problem_2018}, so the hypothesis of temporal closeness is often used in temporal transformations. Figure~\ref{fig: temporal_transformation} shows an example of temporal transformation. A meteorological station recorded weather observations at 9:33 AM, 10:20 AM, 11:08 AM, and 11:53 AM. To estimate the weather at 12:00 PM, we may use the 11:53 AM observation, as it is closest to the target time.
    
    \item \textit{Spatio-temporal transformation:} Spatio-temporal transformations can be regarded as an integration of temporal and spatial transformations, thus adopting the hypothesis of both spatial proximity and temporal closeness.
\end{itemize}

\subsubsection{Dependency Modeling}
\begin{figure}[htbp]
    \centering
    \includegraphics[width=.85\textwidth]{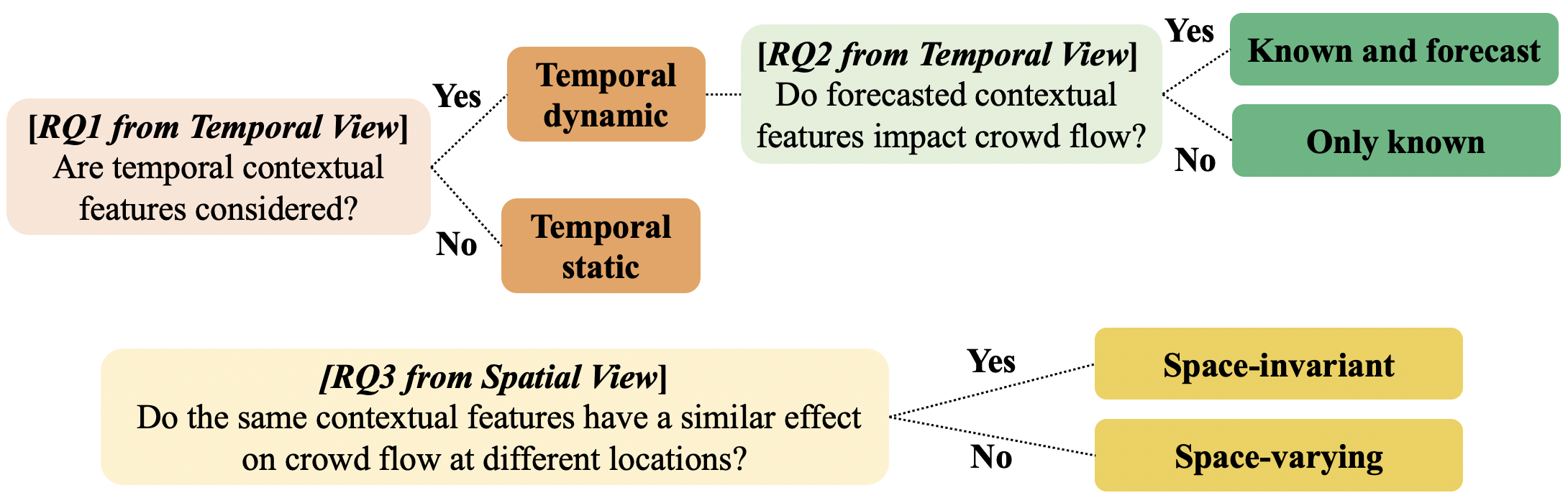}
    % \vspace{.5em}
    \caption{Research questions and hypotheses for modeling context dependency.}
    \label{fig: modeling_gist}
% \vspace{-1em}
\end{figure}

Considering that contextual features have their distinct characteristics (e.g., temporal dynamic for temporal context), dependency modeling aims to learn more effective compressed representations using deep learning techniques (e.g., GNNs for spatial representation \cite{multi_graph_region_embedding_2022}). As shown in Figure~\ref{fig: modeling_gist}, the dependency modeling can be reviewed by answering the following three research questions from temporal and spatial views, respectively.

\begin{itemize}[leftmargin=2em]
\item From the temporal view, we have two research questions to answer: RQ1) \textit{Are temporal contextual features considered?} RQ2) \textit{Do forecasted contextual features impact crowd flow?} If the answer to RQ1 is yes, the \textit{temporal dynamic} hypothesis suggests that the impact of contextual features changes over time, which helps model the temporal heterogeneity (e.g., daily patterns differ between weekdays and weekends \cite{liu_deeppf:_2019, yi2019citytraffic}) of crowd flow \cite{wang_STDM_survey}. If temporal contextual features are considered, the second research question (RQ2) then examines whether forecasted context is included, leading to two hypotheses: i) The \textit{known} hypothesis suggests that only known contexts (e.g., historical weather, scheduled public holidays) affect crowd mobility. ii) The \textit{known \& forecast} hypothesis argues that both known and forecasted contexts influence crowd mobility. This hypothesis is based on the idea that forecasted weather, such as increased wind speed, can lead to better pollutant dispersion compared to previous calm conditions, thus improving air quality. Consequently, forecasted weather is crucial for air quality forecasting \cite{deep_fusion_net_2018} and may further affect crowd mobility.

\item From the spatial view, we have another research question to answer: RQ3) \textit{Do the same contextual features have a similar effect on crowd flow at different locations?} The \textit{space-invariant} hypothesis suggests that context effects are similar across different locations, while the \textit{space-varying} hypothesis indicates that context effects vary by location \cite{zheng_deepstd_2020}. The \textit{space-invariant} hypothesis suggests that context affects all locations similarly \cite{STRN_2021}, as heavy rain may affect different city regions equally. This hypothesis can be implemented by sharing learnable parameters across locations. Conversely, the \textit{space-varying} hypothesis posits that context affects locations unevenly, as rain impacts commercial and residential areas in varying degrees.
\end{itemize}

Notably, as shown in Table~\ref{tab: representative_models}, hypotheses can be considered from either the temporal or spatial view, allowing temporal or spatial context to be applied individually (e.g., \textit{MVGCN} \cite{sunIrregular} uses the \textit{known} hypothesis for temporal context). Additionally, spatial and temporal contexts can be combined to form a spatio-temporal context, as these feature combinations exhibit both temporal dynamics and spatial heterogeneity simultaneously \cite{wang_deepsd_2017, fine_grained_2021}.

\subsubsection{Fusion Techniques} 
In recent years, various feature fusion techniques have been developed to integrate context and crowd flow representations for learning more comprehensive representations \cite{zheng_data_fusion_2015,liu_deep_data_fusion_2020,zou_deep_fusion_2024}. Among them, \textit{feature concatenation} is widely used to fuse features, assuming a low correlation between them to avoid redundancy \cite{context_generalizability}. In contrast, \textit{addition-based methods} assume that crowd flow representations and contextual representations share similar semantics, making them additive in vector space. Recent studies emphasize the effectiveness of \textit{gating mechanisms} \cite{zhang_flow_2019,context_generalizability}, which model the influence of context on crowd flow (e.g., heavy rain reducing bike-sharing usage like a switch) by mapping context into scaling factors that adjust crowd flow representations.

\subsubsection{Training Strategies} 
The goal of training strategies is to optimize three networks: the spatio-temporal crowd flow backbone network for learning mobility data patterns, the context network for capturing context dependencies, and their fusion network. A straightforward training approach is to combine these networks (as shown in Figure~\ref{fig: training_strategies}) and train them together in an \textit{end2end} manner, where all three networks are optimized from scratch using back-propagation. Alternatively, the \textit{pretrain and finetune} strategy first trains the traffic backbone network for prediction tasks, then freezes it while finetuning the context and fusion networks \cite{chai_multi_graph_2018}. This approach may be more effective because traffic patterns are typically more complex than context patterns, resulting in the context network having fewer parameters and requiring less gradient propagation \cite{importance_momentum_2013, LeCun_efficient_back_2012}. Additionally, pretraining and freezing the traffic backbone may help prevent overfitting, as it allows the traffic network to focus solely on learning the traffic patterns, potentially improving generalizability.

\section{Empirical Studies}

In this section, we elaborate on further explorations enabled by the \texttt{STContext} dataset and the proposed context incorporation paradigm.
We first detail the evaluation settings in Section \ref{sec: experiment_setting}. Next, we explore the components of our proposed context incorporation paradigm. While previous studies have examined representation fusion components, such as late and early fusion techniques \cite{zhang2016dnn, lin2019deepstn+}, and recent research has evaluated existing fusion methods through extensive benchmarks \cite{context_generalizability}, our study focuses primarily on the modeling hypothesis and training strategies components in Sections~\ref{sec: analysis_modeling_hypothesis} and \ref{sec: analysis_training_strategies}, respectively. 
Then, we comprehensively evaluate all the contexts we have gathered in Section~\ref{sec: analysis_context_types}. Based on these results, we conclude several important findings.

\subsection{Evaluation Configurations} \label{sec: experiment_setting}
\subsubsection{Datasets} 
\texttt{STContext} includes nine datasets across five typical STCFP tasks. We selected five datasets from different tasks for benchmark experiments to ensure the generalizability of our conclusions. Each dataset covers six months with a time granularity of 60 minutes. We divided the dataset into training, validation, and test sets in chronological order, using the last 20\% of the duration for testing and the 10\% prior to that for validation. Table~\ref{table: benchmark_dataset_stat} presents the statistics for the datasets used in our benchmark experiment.

\begin{table}[htbp]
    \small
    \caption{Statistics of the datasets used in empirical studies. (Wea. for Weather, Demo for Demographics, A.D. for Administrative Division, N/A for not used in empirical studies)}
    \vspace{-0.5em}
    \resizebox{0.93\textwidth}{!}{
    \begin{tabular}{cccccccccc}
    \toprule
\textbf{Datasets} & \textbf{\# Historical Wea.} &  \textbf{\# Wea. Forecast} & \textbf{\# AQI} & \textbf{\# Holiday} & \textbf{\# POI} & \textbf{\# Road} & \textbf{\# Demo} & \textbf{\# A.D.}\\ 
\midrule
Bike\_NYC  & 24,619  & 1,025,037 & 309,660 & 57  & 16,936 & 3,499 & 37,991  & 262 \\
Taxi\_NYC  & 24,619  & 1,025,037 & 309,660 & 57 &  16,936   & 3,499 & 37,991  & 262  \\
Pedestrian\_MEL  & 12,619  & 709,560 & N/A &55 & 26,712  & 6,187 & 11,304 & 17   \\
Speed\_BAY  & 12,376  & 1,935,960 & N/A &  58 & 3,209 & 42,788 & 77,921  & 23 \\
Metro\_NYC  & 14,975 & 1,024,920 & 270,997 & 56  & 49,255 & 29,140 & 37,991  & 262 \\
\bottomrule
    \end{tabular}}
    \label{table: benchmark_dataset_stat}
\end{table}

\subsubsection{Evaluation Metrics} We employ RMSE (Root Mean Square Error) and SMAPE (Symmetric Mean Absolute Percentage Error) as our evaluation metrics as in previous studies \cite{zhang2017deep, zheng_deepstd_2020, dl_traffic_2021, one4all_2024}. 
Previous methods evaluate context-aware STCFP models using a single metric for the entire dataset (e.g., overall RMSE) \cite{zhang2017deep, adaptive_fusion_metro_2023}. However, we argue that contexts like rain, strong winds, and dense fog are less common than clouds or haze, and relying on an overall metric may obscure the model's ability to predict crowd mobility during unusual conditions (e.g., heavy rain). To overcome this, we adopt separate metrics for these atypical moments, as suggested in prior studies \cite{storm_tits_2024}.

Moreover, spatial contexts are heterogeneous (e.g., points of interest in commercial areas typically exceed those in suburban regions). Evaluating them with overall metrics can lead to inaccuracies in assessing the model's effectiveness in predicting crowd mobility across different areas. To address this, we establish spatial divisions for calculating evaluation metrics for each region. Figure \ref{fig: spatial_scenarios_division} shows the divisions for NYC, Melbourne, and the Bay Area, based on administrative boundaries and the volume of spatio-temporal data. Areas with substantial data are recognized as central regions; for example, as shown in Figure \ref{nyc_division}, the Manhattan districts in NYC are identified as central areas.

\newcommand{\subfigcol}{-1mm}
\begin{figure}[htbp]
\begin{minipage}[c]{1\linewidth}
\centering
\subfigure[\textit{NYC}]{
\includegraphics[width=0.3\linewidth]{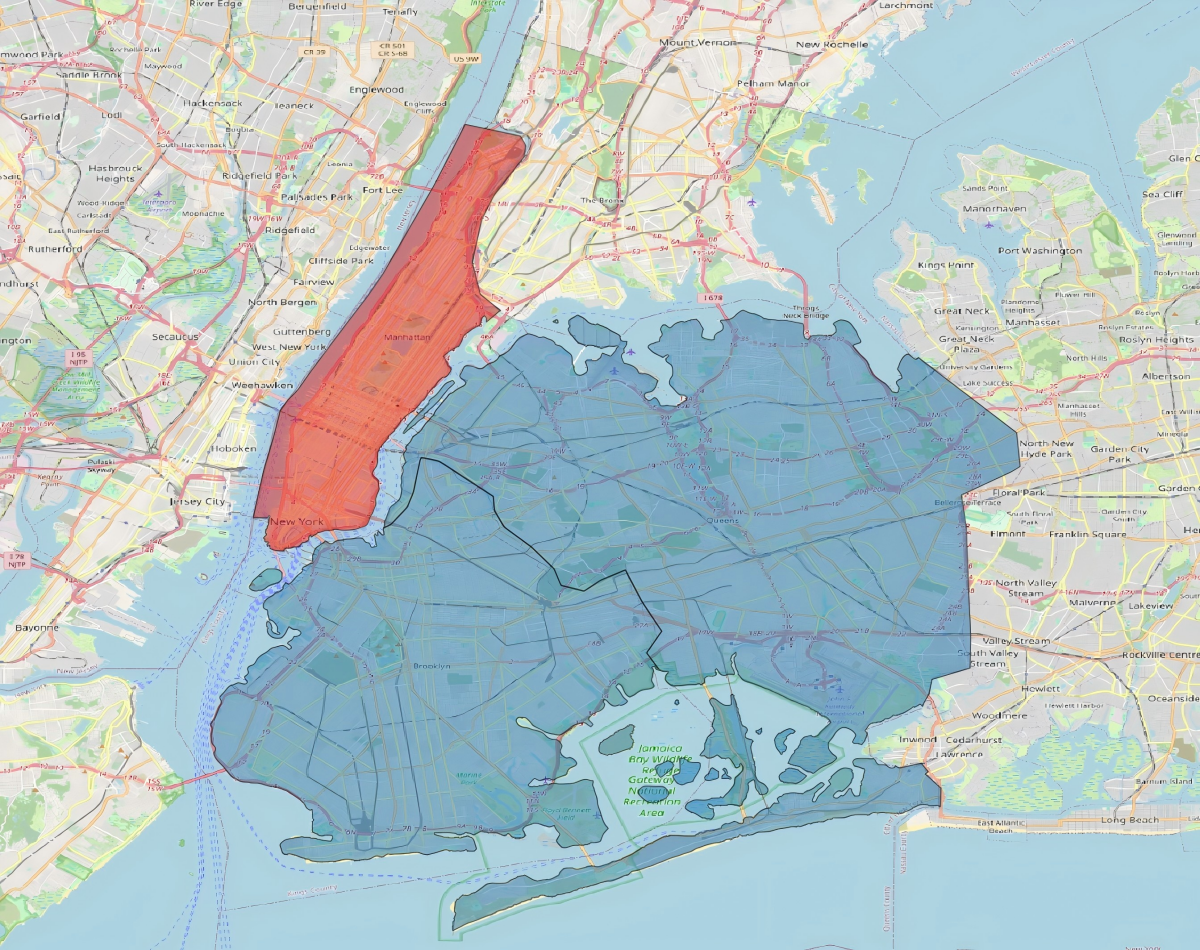}
    \label{nyc_division}
}\hspace{\subfigcol}
\subfigure[\textit{Melbourne}]{
\includegraphics[width=0.3\linewidth]{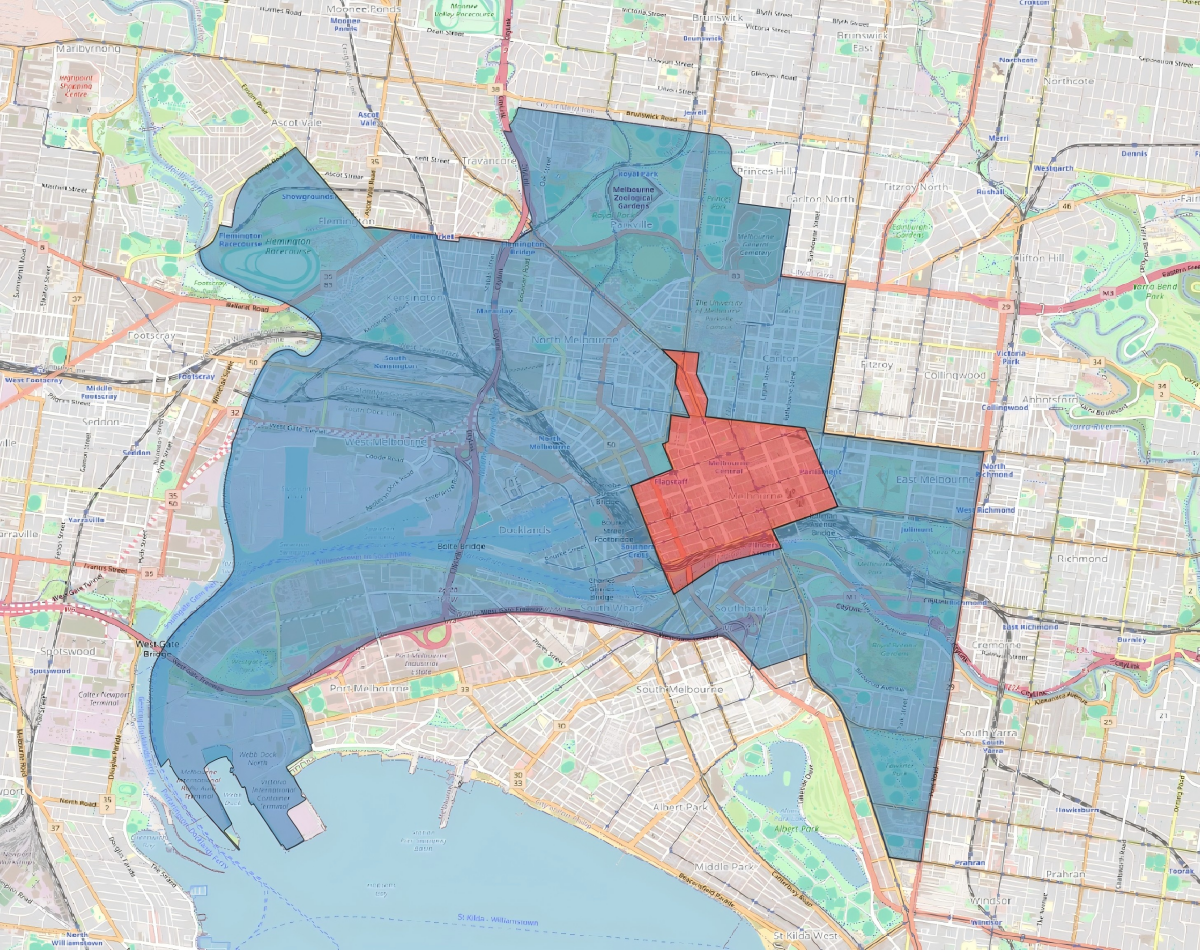}
    \label{melbourne_division}
}\hspace{\subfigcol}
\subfigure[\textit{BAY}]{
\includegraphics[width=0.3\linewidth]{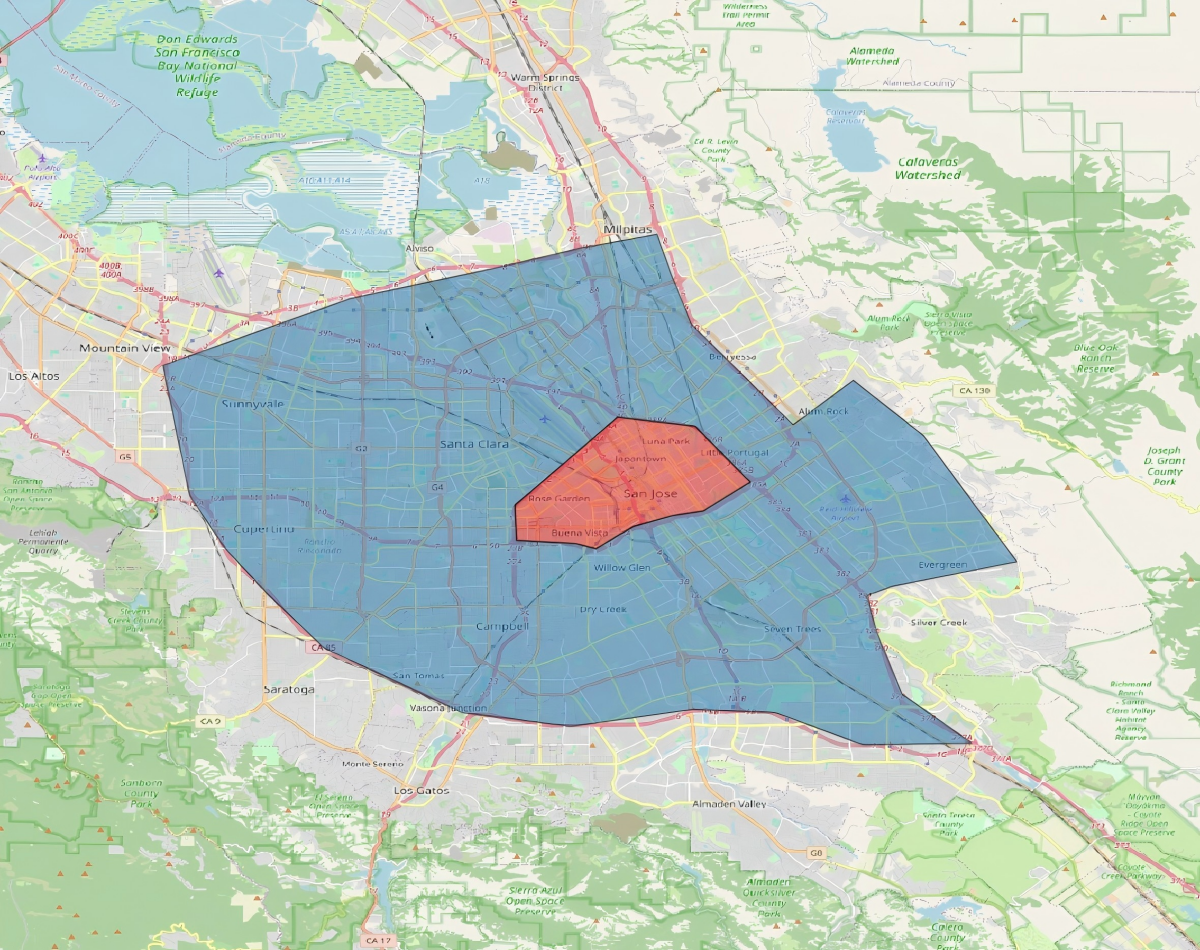}
    \label{bay_division}
}
\vspace{-1em}
\caption{Visualization of spatial divisions in NYC, Melbourne, and the Bay Area. We calculate evaluation metrics for central, non-central, and overall regions, respectively. Central areas are shown in red, while non-central areas are in blue.}
\label{fig: spatial_scenarios_division}
\end{minipage}
\vspace{-.5em}
\end{figure}

\begin{table}[h]
    \centering
    % \small
    \renewcommand{\arraystretch}{1.3}
    \tabcolsep=1mm
    \caption{Results of incorporating weather (spatio-temporal context) under different dependency modeling hypotheses. Variants marked with $^*$ significantly ($p < 0.05$) outperform \textit{No Context}. The best results are in bold.}
    \label{tab: weather_performance}
    \resizebox{.9\textwidth}{!}{
    \begin{tabular}{lc|cc|cc|cc|cc}
\hline
\multirow{2}{*}{\textbf{Dataset}} & \multirow{2}{*}{\textbf{Modeling Hypothesis}} & \multicolumn{2}{c}{\textbf{Overall}} & \multicolumn{2}{c}{\textbf{Rain}} & \multicolumn{2}{c}{\textbf{Wind}} & \multicolumn{2}{c}{\textbf{Fog}} \\ \cline{3-10} 
& & RMSE & SMAPE & RMSE & SMAPE & RMSE & SMAPE & RMSE & SMAPE \\ 
 \hline 
 \multirow{5}{*}{Bike\_NYC}
 & \textit{No Context} & 3.2511 & \textbf{0.1509} & 3.5130 & 0.1964 & 3.4334 & 0.2038 & 2.8402 & 0.1651\\
 & \textit{known \& space-invariant} & 3.2534 & 0.1513 & 3.4831$^*$ & \textbf{0.1952$^*$} & 3.4078$^*$ & 0.2028$^*$ & \textbf{2.7641$^*$} & \textbf{0.1522$^*$}\\
 & \textit{known \& space-varying} & 3.2560 & 0.1521 & 3.4814$^*$ & 0.1968 & 3.4002$^*$ & 0.2037 & 2.7732$^*$ & 0.1536$^*$\\
 & \textit{known and forecast \& space-invariant} & \textbf{3.2425$^*$} & 0.1513 & 3.4731$^*$ & 0.1953$^*$ & 3.3739$^*$ & \textbf{0.2018} $^*$ & 2.8170$^*$ & 0.1542$^*$\\
 & \textit{known and forecast \& space-varying}  & 3.2502 & 0.1526 & \textbf{3.4699$^*$} & 0.1971 & \textbf{3.3657$^*$} & 0.2033 & 2.8265$^*$ & 0.1559$^*$\\
 \hline
 \multirow{5}{*}{Taxi\_NYC}
 & \textit{No Context} & 24.660 & 0.0929 & 26.674 & 0.0898 & 23.046 & 0.0694 & \textbf{29.427} & 0.1024\\
 & \textit{known \& space-invariant} & \textbf{24.572} & 0.0830$^*$ & \textbf{26.655} & 0.0783$^*$ & \textbf{23.004} & \textbf{0.0666} & \textbf{29.427} & 0.0917$^*$\\
 & \textit{known \& space-varying} & 24.573 & \textbf{0.0827$^*$} & 26.662 & \textbf{0.0778$^*$} & 23.025 & \textbf{0.0666} & 29.448 & \textbf{0.0915$^*$}\\
 
 & \textit{known and forecast \& space-invariant} & 25.520 & 0.1119 & 26.771 & 0.1148 & 24.223 & 0.1071 & 29.830 & 0.0972\\
 & \textit{known and forecast \& space-varying} & 24.980 & 0.1126 & 26.979 & 0.1102 & 23.660 & 0.0926 & 30.415 & 0.0964\\
 \hline
 \multirow{5}{*}{Pedestrian\_MEL}
 & \textit{No Context} & 168.94 & 0.2317 & 181.28 & 0.2633 & 220.97 & 0.2533 & 117.77 & 0.2831\\
 & \textit{known \& space-invariant} & 168.86 & 0.2159$^*$ & 179.68$^*$ & 0.2540$^*$ & 219.89 & 0.2435$^*$ & \textbf{116.45$^*$} & 0.2584$^*$\\
 & \textit{known \& space-varying} & 168.84 & 0.2189$^*$ & 179.66$^*$ & 0.2554$^*$ & 219.77 & 0.2447$^*$ & 117.11 & 0.2591$^*$\\
 & \textit{known and forecast \& space-invariant} & \textbf{168.42} &  \textbf{0.2155$^*$} & \textbf{179.62$^*$} & \textbf{0.2539$^*$} & \textbf{219.51$^*$} & \textbf{0.2434$^*$} & 117.04 & 0.2724$^*$\\
 & \textit{known and forecast \& space-varying}  & 168.44 & 0.2158$^*$ & 179.85$^*$ & 0.2549$^*$ & 219.56$^*$ & 0.2439$^*$ & 117.11 & \textbf{0.2548$^*$}\\
 \hline
 \multirow{5}{*}{Speed\_BAY}
 & \textit{No Context} & 3.7505 & 0.0378 & 5.6572 & 0.0744 & 5.1009 & 0.0640 & 4.7654 & 0.0555\\
 & \textit{known \& space-invariant} & 3.7268$^*$ & 0.0367$^*$ & 5.4395$^*$ & 0.0693$^*$ & 5.0820$^*$ & 0.0636 & 4.6189$^*$ & 0.0520$^*$\\
 & \textit{known \& space-varying} & 3.7261$^*$ & 0.0368$^*$ & 5.4442$^*$ & 0.0695$^*$ & 5.0725$^*$ & 0.0634 & 4.6197$^*$ & 0.0521$^*$\\
 & \textit{known and forecast \& space-invariant} & \textbf{3.7041$^*$} & \textbf{0.0362$^*$} & \textbf{5.2779$^*$} & \textbf{0.0653$^*$} & \textbf{5.0111$^*$} & 0.0628$^*$ & \textbf{4.5300$^*$} & \textbf{0.0500$^*$}\\
 & \textit{known and forecast \& space-varying}  & 3.7150$^*$ & 0.0365$^*$ & 5.6416$^*$ & 0.0722$^*$ & 5.0125$^*$ & \textbf{0.0626$^*$ }& 4.7249$^*$ & 0.0531$^*$\\
 \hline
 \multirow{5}{*}{Metro\_NYC}
 & \textit{No Context} & 113.66 & 0.1811 & 129.89 & 0.1644 & 144.84 & 0.1475 & 139.32 & 0.1620\\
 & \textit{known \& space-invariant} & 113.66 & 0.1751$^*$ & 129.77$^*$ & 0.1665 & \textbf{144.33} & 0.1474 & 138.67 & 0.1655\\
 & \textit{known \& space-varying} & \textbf{113.54} & \textbf{0.1645$^*$} & \textbf{129.68} & \textbf{0.1620} & 144.48 & \textbf{0.1448} & 138.86 & \textbf{0.1600}\\
 & \textit{known and forecast \& space-invariant} & 116.32 & 0.1905 & 131.06 & 0.1896 & 145.65 & 0.1971 & \textbf{136.96$^*$} & 0.2002\\
 & \textit{known and forecast} \& \textit{space-varying}  & 119.79 & 0.2325 & 131.61 & 0.2222 & 146.07 & 0.2155 & 139.96 & 0.2245 \\
 \bottomrule
    \end{tabular}}
    % \vspace{-2em}
\end{table}

\subsubsection{Implementation Details} We use MTGNN \cite{MTGNN_2020} as our backbone network to capture the spatio-temporal dependencies of crowd flow, given its strong performance in recent benchmarks \cite{dl_traffic_2021}. We apply the ADAM optimizer with a learning rate of 1e-3. The depth of the mix-hop propagation layer is 2, with a retain ratio of 0.05. The saturation rate of the activation function in the graph learning layer is 3, and the dimensions of node embeddings are 40.

We adopt the proximity hypothesis for spatial transformation, retrieving contextual features within 1.5 km of crowd flow locations, as in previous studies\cite{zheng2015forecasting, deep_fusion_net_2018}. For temporal transformation, we employ the closeness hypothesis, utilizing the contextual features that are nearest in time for each predefined time slot.
To capture temporal dependencies for the \textit{known} and \textit{known \& forecast} hypotheses, we use 2-layer multilayer perceptrons (MLPs), consistent with prior work~\cite{xing2022stgs}. For the \textit{space-invariant} hypothesis, we apply 3-layer MLPs, while for the \textit{space-varying} hypothesis, we implement graph convolutional networks \cite{kipf2017semisupervised}, as in earlier research~\cite{xia_3dgcn_2021}.

Our experimental platform is a server equipped with an Intel(R) Core(TM) i7-11700K CPU @ 3.60 GHz, 64 GB RAM, and an NVIDIA RTX 2080Ti GPU with 11 GB of memory. Each experimental result is validated using five different random seeds, and significance is confirmed through a t-test.

\subsection{Analysis of Modeling Hypotheses} \label{sec: analysis_modeling_hypothesis}
According to our paradigm (see Section~\ref{sec: analysis_modeling_hypothesis}), we can adopt \textit{known} or \textit{known \& forecast} hypotheses to model temporal dependencies, while \textit{space-invariant} and \textit{space-varying} model spatial dependencies.
To evaluate the effectiveness of these modeling hypotheses, we conduct experiments using spatial context (i.e., POI), temporal context (i.e., temporal position), and spatio-temporal context (i.e., weather). We selected these contextual features based on their proven effectiveness in previous studies \cite{tong_simpler_2017, LiBikeTKDE2019, yuan_demand_2021}. 
Table~\ref{tab: weather_performance} displays the results of incorporating weather (spatio-temporal context) across four modeling hypotheses, while Table~\ref{tab: tp_poi_modeling_results} shows the results for POI (spatial context), temporal position (temporal context), and their combination across various modeling hypotheses. From these results, we conclude the following findings.

\subsubsection{Scenario-specific metrics are better suited than overall metrics for evaluating context-aware STCFP models} \label{sec: evaluation_on_specific_metrics}

As shown in Table \ref{tab: weather_performance}, the \textit{known and forecast \& space-invariant} variant in the Bike\_NYC dataset, while significantly outperforming \textit{No Context}, shows minimal overall improvement (less than 0.3\% in RMSE), suggesting that weather features may have limited impact on crowd mobility. 
However, in specific scenarios (e.g., rain, wind, and fog), it achieves over a 2.7\% improvement compared to \textit{No Context}. This discrepancy arises because samples associated with significant rain, wind, and fog represent only around 8\%, 10\%, and 5\% of the total samples, causing overall indicators to dilute the model's impact on these instances. Therefore, to comprehensively and accurately assess the performance of context-aware STCFP models, it is essential to construct appropriate evaluation scenarios. We appeal to design scenario-specific evaluations based on the context characteristic itself. For example, considering that the aim of taking weather into STCFP models is to enhance the prediction under extreme weather, which is rare in real-world scenarios, evaluating models in such atypical weather states is a reasonable way.

\begin{table}[htbp]
    \caption{Results of POI (spatial context), temporal position (temporal context), and TP$\times$POI (the combination of TP and POI, recognized as spatio-temporal context) under different dependency modeling hypotheses. Variants marked with $^*$ significantly ($p < 0.05$) outperform \textit{No Context}. The best results are in bold.}
    %     \small
    % \renewcommand{\arraystretch}{1.3}
    % \tabcolsep=1mm
    \label{tab: tp_poi_modeling_results}
\resizebox{.9\textwidth}{!}{
\begin{tabular}{lc|cc|cc|cc}

\hline  \multirow{2}{*}{\textbf{Dataset}} & \multirow{2}{*}{\textbf{Modeling Hypothesis}}& \multicolumn{2}{c}{ \textbf{Overall}} & \multicolumn{2}{c}{\textbf{Central}} & \multicolumn{2}{c}{\textbf{Non-central}}   \\ \cline{3-8}
 & & RMSE & SMAPE & RMSE & SMAPE &RMSE & SMAPE  \\ 
\hline \multirow{5}{*}{Bike\_NYC}& \textit{No Context} &3.2511 & 0.1509 & 4.0402 & 0.3492 & 1.7483 & \textbf{0.2658} \\
 & \textit{space-invariant} (\textit{POI})&3.1889$^*$ &0.1472 & 3.9852$^*$ & 0.3477$^*$ & 1.7334$^*$ & 0.2742 \\
 & \textit{space-varying} (\textit{POI}) & 3.1908$^*$&0.1477 & 3.9875$^*$ & 0.3484 & 1.7343$^*$ & 0.2747 \\
 & \textit{known} (\textit{TP}) &3.0700$^*$& 0.1422$^*$ & 3.8267$^*$&0.3381$^*$ &1.6971$^*$& 0.2728 \\
 & \textit{known \& space-invariant} (\textit{TP $\times$ POI}) & \textbf{3.0569$^*$} &\textbf{0.1421$^*$} &\textbf{3.8078$^*$} &\textbf{0.3374$^*$} &\textbf{1.6969$^*$} &0.2761 \\
\hline 

\multirow{5}{*}{Taxi\_NYC }& \textit{No Context} & 24.660 &0.0929 & 42.732 & 0.1818& 9.9874 & 0.0276    \\
 & \textit{space-invariant} (\textit{POI})& 24.087$^*$& 0.0803$^*$& 42.042$^*$ & 0.1613$^*$ & \textbf{9.7171$^*$} & \textbf{0.0251$^*$} \\
 & \textit{space-varying} (\textit{POI})& 24.099$^*$&\textbf{0.0801$^*$} & 42.016$^*$ & 0.1625$^*$ & 9.8061$^*$ & 0.0256$^*$ \\
 & \textit{known} (\textit{TP})& 22.329$^*$ &0.0940 & 38.297$^*$ &\textbf{0.1469$^*$} &10.123 &0.0270 \\
 & \textit{known \& space-invariant} (\textit{TP $\times$ POI}) &\textbf{22.279$^*$}&0.0809$^*$&\textbf{38.115$^*$}& 0.1472$^*$ & 10.249 & 0.0266 \\

\hline 
\multirow{5}{*}{Pedestrian\_MEL } & \textit{No Context} & 168.94 & 0.2317 & 128.51 & 0.2832 & 219.12 & 0.3187  \\
 & \textit{space-invariant} (\textit{POI}) &165.47$^*$ &0.2239$^*$& 127.64$^*$ & 0.2662$^*$ & 219.98 & \textbf{0.2777$^*$}  \\
 & \textit{space-varying} (\textit{POI})& 165.40& 0.2328 & 127.40$^*$ & 0.2860 & 220.08 & 0.3279 \\
 & \textit{known} (\textit{TP}) &152.78$^*$ &0.2250$^*$ &110.56$^*$ & 0.2571$^*$ &210.67$^*$ & 0.2856$^*$ \\
& \textit{known \& space-invariant} (\textit{TP $\times$ POI})& \textbf{152.20$^*$}&\textbf{0.2225$^*$} &\textbf{110.07$^*$} & \textbf{0.2533$^*$} & \textbf{209.99$^*$} &0.2842$^*$ \\

\hline
\multirow{5}{*}{Speed\_BAY} & \textit{No Context}& 3.5321 & 0.0340 & 3.4606 & 0.0330 & 3.5375 & 0.0341 \\
 & \textit{space-invariant} (\textit{POI}) & 3.4790& 0.0336& 3.4232 & 0.0326& 3.4833 & 0.0338 \\
 & \textit{space-varying} (\textit{POI})& 3.4791& 0.0336 & 3.4230 & 0.0324 & 3.4833 & 0.0336 \\
 & \textit{known} (\textit{TP}) &3.3882$^*$ & 0.0325$^*$ & 3.3298$^*$ & 0.0313$^*$&3.3927$^*$ &  \textbf{0.0325$^*$}  \\
 & \textit{known \& space-invariant} (\textit{TP $\times$ POI}) & \textbf{3.3810}$^*$& \textbf{0.0324$^*$} & \textbf{3.3217$^*$} & \textbf{0.0312$^*$} & \textbf{3.3855$^*$} & \textbf{0.0325$^*$} \\

\hline 
\multirow{5}{*}{Metro\_NYC} & \textit{No Context}  & 105.90 & 0.1664  & 155.27 & 0.2131 &78.487 & 0.2254  \\
 & \textit{space-invariant} (\textit{POI})& 105.08$^*$& 0.1650$^*$ & 154.05$^*$ & 0.2161 & 77.878$^*$ & 0.2136$^*$ \\
& \textit{space-varying} (\textit{POI})& 104.97$^*$&0.1648$^*$ & 154.00$^*$ & 0.2110$^*$ & 77.721$^*$ & 0.2141$^*$  \\
 & \textit{known} (\textit{TP}) &97.061$^*$ & 0.1618 & 135.74$^*$ &0.2153 &76.731$^*$ & 0.2350 \\
 & \textit{known \& space-invariant} (\textit{TP $\times$ POI})& \textbf{96.608}$^*$ & \textbf{0.1574}$^*$&\textbf{135.10$^*$} &\textbf{0.2039$^*$} & \textbf{76.386$^*$} &\textbf{0.2052$^*$} \\
\bottomrule
\end{tabular}}
\vspace{-1em}
\end{table}

\subsubsection{The \textit{space-varying} modeling hypothesis, while taking more parameters, does not outperform the \textit{space-invariant} hypothesis} 

As shown in Table~\ref{tab: weather_performance} and Table~\ref{tab: tp_poi_modeling_results}, \textit{space-varying} variants, which assume that context affects each location differently (e.g., heavy rain impacting commercial and residential areas unequally), generally do not outperform \textit{space-invariant} variants across most tasks. 
Even in both central and non-central areas, \textit{space-invariant} variants perform similarly to \textit{space-varying} variants, indicating that both hypotheses may learn comparable context representations.
This outcome is somewhat surprising, as \textit{space-varying} variants have more parameters and were expected to perform better \cite{zheng_deepstd_2020}.
The inconsistent result may stem from two issues. First, the \textit{space-invariant} hypothesis may be inherently more robust, suggesting that contextual features like weather and POI often exert similar effects across urban areas. Second, the added complexity of the \textit{space-varying} hypothesis makes it harder to optimize, increasing the risk of overfitting.

\subsubsection{Combining effective spatial and temporal dependency modeling leads to better predictions than using either alone}

Table \ref{tab: tp_poi_modeling_results} shows that applying the \textit{space-invariant} and \textit{known} hypotheses to POI and temporal position, respectively, significantly outperforms the \textit{No Context} baseline, confirming their effectiveness and aligning with prior studies \cite{STRN_2021, MVSTGN_2023_TMC}. We observe that integrating temporal position with the \textit{known} hypothesis produces a greater improvement (e.g., over 9\% RMSE reduction in Taxi\_NYC) than incorporating POI through the \textit{space-invariant} hypothesis, which is also consistent with past findings \cite{context_generalizability}.
As these hypotheses independently learn context from time and space, the \textit{known \& space-invariant} combination merges them through concatenation, forming a spatio-temporal context that varies across time and space. This combined variant generally surpasses the performance of either \textit{known} or \textit{space-invariant} alone, as shown in Table \ref{tab: tp_poi_modeling_results}.
Moreover, the greater the improvement of the \textit{space-invariant (POI)} variant over \textit{No Context}, the more the combined variants outperform \textit{known (TP)} (e.g., in the Bike\_NYC dataset), suggesting that spatial and temporal dependency improvements may be independent. 
This insight leads to two key insights: (1) combining effective temporal and spatial contexts further enhances predictions, and (2) advances in either temporal or spatial dependency modeling will improve overall performance.

\subsubsection{The spatial modeling hypothesis applied to POI generally performs better in central areas than in non-central areas.}

As shown in Table \ref{tab: tp_poi_modeling_results}, the spatial modeling hypotheses (i.e., \textit{space-invariant (POI)} and \textit{space-varying (POI)}) generally outperform \textit{No Context}, achieving over 1.8\% RMSE improvement in Bike\_NYC. 
This demonstrates the effectiveness of both POI and the \textit{space-invariant} modeling hypothesis.
More specifically, prediction improvement is generally greater in central areas than in non-central areas, indicating that the modeling hypothesis learns more effective POI representations in locations with much greater crowd flow.
This phenomenon may stem from data quality bias, as central areas represent urban cores with high density and diversity of facilities (e.g., business centers, transportation hubs) and increased crowd flow. These regions often display greater diversity in spatial data, with a higher number and variety of points of interest \cite{yuan_functions_2012}.
In contrast, non-central areas, such as suburbs, have more dispersed data points and a narrower variety of POI types. The data richness in central areas likely enhances model accuracy by providing more robust samples and distinctive features, which helps in distinguishing regions with different functions. 
Hence, it is of great importance to verify spatial data quality when applying spatial modeling hypotheses.

\subsubsection{Accurate forecast data helps enhance prediction performance}\label{sec: real_past_comparison}

\begin{figure}[h]
    \centering
\begin{minipage}{0.43\textwidth}
    \centering
    \includegraphics[width=1\textwidth]{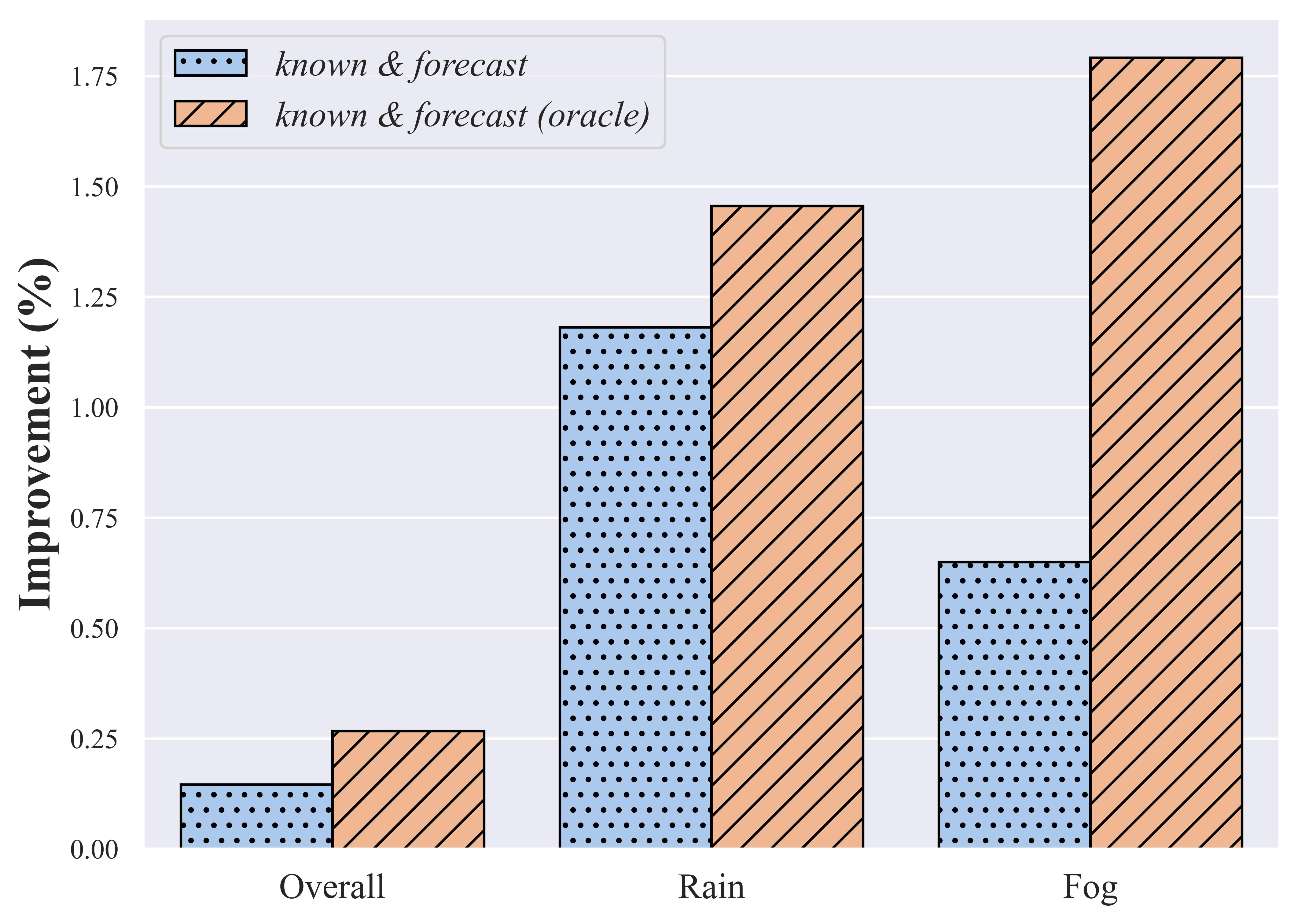}
    % \vspace{-2em}
    \caption{\textit{known and forecast} vs. \textit{known and forecast (oracle)}. As time progresses, the forecast for time $t$ is replaced by accurate real data collected at time $t$, making a variant called \textit{forecast (oracle)}.}
    % \vspace{-1em}
    \label{fig: forecast_vs_historical}
\end{minipage}%
\hspace{1em}
\begin{minipage}{0.53\textwidth}
\centering
    \small
    \captionof{table}{Results of different training strategies. Variants marked with $^*$ significantly ($p < 0.05$) outperform \textit{No Context}. The best results are in bold.}
    \vspace{-1em}
    \label{tab: training_strategies_results}
    \resizebox{.9\textwidth}{!}{
    \begin{tabular}{lc|ccc}
\hline
\multirow{2}{*}{\textbf{Dataset}} & \multirow{2}{*}{\textbf{Training Strategy}} & \multicolumn{2}{c}{\textbf{Overall}}  \\ \cline{3-4} 
& & RMSE & SMAPE \\ 
\hline 
\multirow{3}{*}{Bike\_NYC}
& \textit{No Context} & 3.2511 & 0.1509 \\
& \textit{End2end} & 3.0770$^*$ & 0.1459$^*$ \\
& \textit{Pretrain \& Finetune} & \textbf{3.0569$^*$} & \textbf{0.1421$^*$} \\
\hline

\multirow{3}{*}{Taxi\_NYC}
& \textit{No Context} & 24.660 & 0.0929 \\
& \textit{End2end} & 22.431$^*$ & 0.0870$^*$ \\
& \textit{Pretrain \& Finetune} & \textbf{22.279$^*$} & \textbf{0.0809$^*$} \\
\hline

\multirow{3}{*}{Pedestrian\_MEL}
& \textit{No Context} & 168.94 & 0.2317 \\
& \textit{End2end} & 153.01$^*$ & 0.2257$^*$ \\
& \textit{Pretrain \& Finetune} & \textbf{152.20$^*$} & \textbf{0.2225$^*$} \\
\hline

\multirow{3}{*}{Speed\_BAY}
& \textit{No Context} & 3.5321 & 0.0340 \\
& \textit{End2end} & 3.3835$^*$ & \textbf{0.0324$^*$} \\
& \textit{Pretrain \& Finetune} & \textbf{3.3810$^*$} & \textbf{0.0324$^*$} \\
\hline

\multirow{3}{*}{Metro\_NYC}
& \textit{No Context} & 105.90 & 0.1664 \\
& \textit{End2end} & 97.645$^*$ & 0.1696$^*$ \\
& \textit{Pretrain \& Finetune} & \textbf{96.608$^*$} & \textbf{0.1574$^*$} \\
\hline
\end{tabular}}  
\end{minipage}
\end{figure}

As shown in Table~\ref{tab: weather_performance}, the \textit{known and forecast} variants outperform the \textit{known} variants in Bike\_NYC, indicating that forecast information is beneficial for bike flow prediction. However, forecast data can often be inaccurate (e.g., weather forecasts could vary from actual conditions). To explore the impact of more accurate forecast data, we introduce a new variant, \textit{known and forecast (oracle)}, which replaces forecasted weather with actual data collected later.
The experimental results in Figure~\ref{fig: forecast_vs_historical} (with the vertical axis showing improvement over \textit{No Context}) reveal that the \textit{forecast (oracle)} variant significantly improves performance in the overall, rain, and fog scenarios. It suggests that forecast accuracy is likely a crucial factor in limiting the \textit{forecast} hypothesis for further enhancing prediction performance.

\subsection{Analysis of Training Strategies} \label{sec: analysis_training_strategies}
To evaluate the effectiveness of different training strategies, we conduct experiments comparing \textit{end2end} training with \textit{pretrain and finetune (PT\&FT)} across five datasets, with results shown in Table \ref{tab: training_strategies_results}. We utilize the combination of temporal position and POI, which proved effective in the previous experiments (see Table~\ref{tab: tp_poi_modeling_results}). From the results, we make the following observations.

\subsubsection{Both \textit{end2end} training and \textit{PT\&FT} are effective strategies for learning context representations.}
Table~\ref{tab: training_strategies_results} shows that both \textit{end2end} and \textit{PT \& FT} significantly outperform \textit{No Context} ($p < 0.05$), indicating that both strategies effectively learn crowd flow and context representations for prediction. 
This aligns with prior research findings \cite{zheng_deepstd_2020, STRN_2021, MVSTGN_2023_TMC} showing strong results with \textit{end2end} training. Furthermore, while \textit{PT \& FT} generally outperforms \textit{end2end}, as later discussed, their RMSE and SMAPE metrics differ by less than 1\%. Thus, compared to decisions around contextual feature selection or modeling hypotheses, the impact of choosing between these training strategies is relatively minor.

\subsubsection{\textit{PT\&FT} is generally better than end2end training in learning context representations.}
As shown in Table~\ref{tab: training_strategies_results}, \textit{PT\&FT} and \textit{end2end} training perform similarly across all datasets in RMSE and SMAPE, although small differences exist. Specifically, \textit{PT\&FT} generally outperforms \textit{end2end} in most datasets, with improvements of up to 0.66\% in terms of RMSE, suggesting it learns slightly better representation for predictions. 
This aligns with our expectations, as crowd flow patterns are often more complex than context patterns, resulting in crowd flow backbone networks having more parameters \cite{zhang_flow_2019, STRN_2021}. In \textit{end2end} training, we simultaneously optimize context and crowd flow networks through gradient backpropagation. However, context networks may converge before crowd flow networks do, causing ongoing gradients from \textit{end2end} training to update context networks, which increases the risk of overfitting and may degrade generalizability \cite{importance_momentum_2013, LeCun_efficient_back_2012}.

\subsection{Analysis of Contextual Features} \label{sec: analysis_context_types}

\begin{table}[htbp]
    \centering
    \small
    \tabcolsep=5mm
    \caption{Results of AQI (spatio-temporal context) under different modeling hypotheses. As forecast AQI data is unavailable, we conduct experiments using only the \textit{known} hypothesis. No variants significantly ($p < 0.05$) outperform \textit{No Context}. The best results are in bold.}
    \label{tab: AQI_ST_results}
    % \resizebox{.6\textwidth}{!}{
    \begin{tabular}{lc|cc}
\hline
\multirow{2}{*}{\textbf{Dataset}} & \multirow{2}{*}{\textbf{Modeling Hypothesis}} & \multicolumn{2}{c}{\textbf{Overall}}  \\ \cline{3-4} 
& & RMSE & SMAPE  \\ 
 \hline
 \multirow{3}{*}{Bike\_NYC}
 & \textit{No Context} & \textbf{3.2511} & \textbf{0.1509} \\
 & \textit{known \& space-invariant} & \textbf{3.2511} & 0.1512 \\
 & \textit{known \& space-varying} & 3.2538 & 0.1514 \\
 \hline
 \multirow{3}{*}{Taxi\_NYC}
 & \textit{No Context} & 24.660 & 0.0929 \\
 & \textit{known \& space-invariant} & 24.571 & 0.0935 \\
 & \textit{known \& space-varying} & \textbf{24.566} & \textbf{0.0928} \\
 \hline
 \multirow{3}{*}{Metro\_NYC}
 & \textit{No Context} & 113.66 & \textbf{0.1811} \\
 & \textit{known \& space-invariant} & 113.59 & 0.1861 \\
 & \textit{known \& space-varying} & \textbf{113.58} & 0.1845 \\
 \bottomrule
    \end{tabular}
    % \vspace{-2em}
\end{table}

\subsubsection{AQI is less effective for enhancing crowd mobility prediction, likely due to people's high tolerance for air pollution in their mobility patterns}

To explore the impact of different types of spatio-temporal context on crowd flow prediction, we conducted experiments using AQI in the Bike\_NYC, Taxi\_NYC, and Metro\_NYC datasets. Since forecast AQI data is unavailable, we model context dependencies using \textit{known} variants.
The results in Table~\ref{tab: AQI_ST_results} show that AQI contributes minimally across all tasks (yielding less than a 0.1\% improvement compared to \textit{No Context}), indicating that AQI is not effective for enhancing crowd mobility predictions.
This finding is also supported by Figure~\ref{fig: pm25_correlation}, which indicates that air pollution has less impact on both bike and taxi flows, further revealing that people tend to tolerate air pollution in their mobility patterns~\cite{chen2021aqi_impact,xia2024escaping}. Furthermore, the above analysis raises an interesting question: \textit{can we find beneficial contextual feature sets for a specific mobility prediction task?} While a straightforward approach is to conduct an ablation study on contextual features, this requires extensive experiments. Instead, the analysis suggests a more efficient method—using data analysis to assess feature effectiveness with less effort. A promising direction is to develop a data analysis framework for selecting contextual features in any given prediction task.

\begin{table}[htbp]
    \centering
    \small
    \caption{Results for various spatial contexts (i.e., POI, administrative diversion, demographics, and road) under the \textit{known \& space-invariant} hypothesis. Variants with $^*$ significantly outperform \textit{No Context} ($p < 0.05$). The best results are in bold.}
    \label{tab: different_spatial_feature_result}
    % \resizebox{.5\textwidth}{!}{
    \begin{tabular}{lc|cc|cc|cccc}

\hline  \multirow{2}{*}{\textbf{Dataset}} & \multirow{2}{*}{\textbf{Feature Type}}& \multicolumn{2}{c}{ \textbf{Overall}} & \multicolumn{2}{c}{\textbf{Central}} & \multicolumn{2}{c}{\textbf{Non-central}}   \\ \cline{3-8}
 & & RMSE & SMAPE & RMSE & SMAPE &RMSE & SMAPE  \\ 

\hline 
\multirow{6}{*}{Bike\_NYC}
& \textit{No Context} &3.2511 & 0.1509 & 4.0402 & 0.3492 & 1.7483 & \textbf{0.2658}  \\
& \textit{TP} &3.0700$^*$ & 0.1422$^*$ & 3.8267$^*$ & 0.3381$^*$ & 1.6971$^*$ &0.2728 \\
& \textit{TP $\times$POI} & \textbf{3.0569$^*$} & 0.1421$^*$ &\textbf{3.8078$^*$} &0.3374$^*$ &1.6969$^*$ &0.2761 \\
& \textit{TP $\times$A.D.} & 3.0669$^*$ &0.1419$^*$&3.8217$^*$ &0.3384$^*$ &1.6985$^*$ &0.2667 \\
& \textit{TP $\times$Demographics} & 3.0940$^*$ & 0.1431$^*$ &3.8578$^*$ & 0.3414$^*$ &1.7072$^*$ &0.2750  \\
& \textit{TP $\times$Road} & 3.0596$^*$ &  \textbf{0.1418$^*$} &3.8133$^*$ &\textbf{0.3364}$^*$ &\textbf{1.6926$^*$} &0.2718  \\
\hline
\multirow{6}{*}{Taxi\_NYC}
& \textit{No Context} & 24.660 &0.0929 &  42.732 & 0.1818& \textbf{9.9874} & 0.0276    \\
& \textit{TP} &22.329$^*$ & 0.0940 & 38.297$^*$ & 0.1469$^*$ & 10.123 & 0.0270 \\
& \textit{TP $\times$POI}& 22.279$^*$& 0.0809$^*$&38.115$^*$& 0.1472$^*$ & 10.249 & 0.0266 \\
& \textit{TP $\times$A.D.} & \textbf{22.151}$^*$&  \textbf{0.0794}$^*$&\textbf{38.027}$^*$ &0.1494$^*$ &9.9896 &\textbf{0.0255} \\
& \textit{TP $\times$Demographics} & 22.434$^*$ &  0.0832$^*$&38.593$^*$ & 0.1527$^*$& 9.9949& 0.0268   \\
& \textit{TP $\times$Road} & 22.379$^*$ & 0.0804$^*$ &38.445$^*$ &\textbf{0.1459}$^*$ &10.051 & 0.0256 \\

\hline

\multirow{6}{*}{Pedestrian\_MEL}
& \textit{No Context} & 168.94 & 0.2317 & 128.51 & 0.2832 & 219.12 & 0.3187  \\
& \textit{TP} &152.78$^*$ & 0.2250$^*$ & 110.56$^*$ & 0.2571$^*$ & 210.67$^*$ & 0.2856$^*$ \\
& \textit{TP $\times$POI}  &152.20$^*$&0.2225$^*$ &110.07$^*$ &  \textbf{0.2533$^*$} & 209.99$^*$ &0.2842$^*$ \\
& \textit{TP $\times$A.D.}& 152.64$^*$ & 0.2179$^*$ &110.65$^*$ &0.2569$^*$ &210.34$^*$ &0.2819$^*$ \\
& \textit{TP $\times$Demographics} &\textbf{149.28}$^*$ &\textbf{ 0.2124} & 112.43$^*$&0.2590$^*$ & \textbf{201.38}$^*$& \textbf{0.2805 }$^*$ \\
& \textit{TP $\times$Road} & 152.39$^*$ & 0.2285$^*$ &\textbf{109.74}$^*$  &0.2655$^*$ &210.72$^*$ & 0.3021$^*$ \\

\hline
\multirow{6}{*}{Speed\_BAY}
& \textit{No Context}& 3.5321 & 0.0340 & 3.4606 & 0.0330 & 3.5375 & 0.0341 \\
& \textit{TP} &3.3882$^*$ & 0.0325$^*$ & 3.3298$^*$ & 0.0313$^*$ & 3.3927$^*$ & \textbf{0.0325$^*$} \\
& \textit{TP $\times$POI}  & \textbf{3.3810$^*$}& \textbf{0.0324$^*$} & \textbf{3.3217$^*$} & \textbf{0.0312$^*$} & \textbf{3.3855$^*$} & \textbf{0.0325$^*$} \\
& \textit{TP $\times$A.D.}&  3.3880$^*$ & \textbf{0.0324$^*$} & 3.3317$^*$&0.0313$^*$ & 3.3923$^*$ &\textbf{0.0325$^*$}  \\
& \textit{TP $\times$Demographics} & 3.3927$^*$& 0.0325$^*$ &3.3446$^*$ &0.0316$^*$& 3.3964$^*$ & 0.0326$^*$  \\
& \textit{TP $\times$Road} & 3.3898$^*$ & 0.0325$^*$ &3.3364$^*$ &0.0315$^*$ &3.3925$^*$ &\textbf{0.0325$^*$} \\

\hline
\multirow{6}{*}{Metro\_NYC}
& \textit{No Context} & 105.90 & 0.1664  & 155.27 & 0.2131 &78.487 & 0.2254 \\
& \textit{TP} & 97.061$^*$ & 0.1618 & 135.74$^*$ & 0.2153 & 76.731$^*$ & 0.2350 \\
& \textit{TP $\times$POI} &96.608$^*$ & \textbf{0.1574}$^*$&\textbf{135.10$^*$} &\textbf{0.2039$^*$} & 76.386$^*$ &\textbf{0.2052$^*$}  \\
& \textit{TP $\times$A.D.} & \textbf{96.534}$^*$& 0.1712 & 135.27$^*$ &0.2115$^*$ &76.204$^*$ &0.2110$^*$ \\
& \textit{TP $\times$Demographics}& 100.28$^*$ & 0.1654$^*$ &144.72 $^*$&  0.2118$^*$& \textbf{76.061}$^*$ &0.2120$^*$  \\
& \textit{TP $\times$Road} & 97.221$^*$ & 0.1698&136.06$^*$ & 0.2122$^*$ & 76.791$^*$&0.2260  \\

\bottomrule
    \end{tabular}
    % \vspace{-2em}
\end{table}

\subsubsection{Spatial contextual features can enhance crowd mobility prediction, but their effectiveness varies by task}

To compare the effectiveness of various spatial contexts, we conducted experiments on five datasets using different spatial context types (POI, Administrative Division, Demographics, and Road) combined with temporal position. We select this combination due to its strong performance with POI in Table~\ref{tab: tp_poi_modeling_results} under the \textit{known \& space-invariant} hypothesis.
Table~\ref{tab: different_spatial_feature_result} shows that no single type of spatial contextual feature consistently outperforms others across all datasets, suggesting that spatial context enhances crowd mobility prediction but varies by task. For example, \textit{TP $\times$ A.D.} performs best in Taxi\_NYC, while \textit{TP $\times$ Demographics} leads in Pedestrian\_MEL.
Moreover, differences among spatial context types are generally minimal, with the largest observed difference being only 1.2\% in RMSE in the Taxi\_NYC dataset.

\subsubsection{Fast food, restaurants, and cafes are beneficial POI types with good generalizability for crowd mobility prediction}

To evaluate the impact of various types of POI on crowd flow prediction, we conducted ablation experiments on the Taxi\_NYC, Metro\_NYC, and Pedestrian\_MEL datasets. We first trained a baseline model including all POI types, and we set each POI feature input to zero individually to observe performance changes. Greater performance degradation indicates higher importance of that POI type.
Table~\ref{tab: types_of_poi} lists the top five most important POI types across all datasets. 
Notably, fast food, restaurants, and cafes enhance prediction performance across tasks, showing their effectiveness and generalizability.
As revealed in previous research~\cite{zhao2023causal}, regional attributes have a strong causal relationship with human mobility, with dining areas attracting people at mealtimes. These findings can guide POI selection for crowd flow prediction and support the potential transferability of models across diverse tasks.

\begin{table}[htbp]
    \caption{Top five most important POI types in Taxi\_NYC, Metro\_NYC, and Pedestrian\_MEL datasets. POI's importance is determined by feature ablation studies.}
    \small
    \renewcommand{\arraystretch}{1.3}
    \tabcolsep=3mm
    \label{tab: types_of_poi}
% \resizebox{.95\textwidth}{!}{
\begin{tabular}{lccccccc}

\toprule  
\multirow{1}{*}{\textbf{Dataset}} & \multirow{1}{*}{\textbf{Most Important POI Types (\textit{Top-5})}} \\
\midrule 
\multirow{1}{*}{Taxi\_NYC} & Fast food, Restaurant, Bank, Cafe, Bicycle rental \\  

\midrule 
\multirow{1}{*}{Metro\_NYC} & Fast food, Restaurant, Cafe, Bicycle parking, School\\  

\midrule 
\multirow{1}{*}{Pedestrian\_MEL} & Fast food, Restaurant, Cafe, Bench, Pub \\  

\bottomrule
\end{tabular}
% \vspace{-1em}
\end{table}

\section{Discussion}
\subsection{Potential Research and Applications of the \texttt{STContext} Dataset}
We argue that the \texttt{STContext} dataset opens up diverse research directions for machine learning scientists and urban transportation researchers. Machine learning scientists may leverage this dataset to develop robust crowd flow prediction models by integrating various contextual factors. That is, building powerful general models that fully utilize diverse contextual features, which help improve the accuracy of crowd mobility predictions. By training models on this comprehensive dataset, researchers may create context-aware one-for-all foundation models for crowd flow prediction in diverse spatio-temporal tasks. Additionally, the \texttt{STContext} dataset presents a challenging setting for machine learning techniques such as few-shot learning \cite{wang2020generalizing} where extreme weather conditions, despite their rarity, significantly impact crowd mobility.

For urban transportation researchers, the \texttt{STContext} dataset aids in understanding how different contextual factors influence mobility patterns. For example, analyzing the effects of holidays or extreme weather on crowd flow can inform infrastructure development and policy decisions. Insights gained from these relationships can enhance crowd mobility simulations, leading to the formulation of improved urban policies aimed at enhancing public transit and reducing traffic accidents \cite{deeptransport_2016, traffic_simulation_2022}.

\subsection{Limitations \& Future Work}

\subsubsection{Limitations}
We primarily collect publicly available data due to licensing constraints, which may result in lower quality compared to commercial sources. For instance, our POI data from OpenStreetMap has several limitations: some areas lack sufficient POI coverage, user tagging can be inconsistent, updates may be delayed, and geographic accuracy can be questionable. Many POIs also lack detailed information, such as operating hours or contact details. In contrast, Google Maps, which charges based on API calls, offers more comprehensive coverage by leveraging user-generated content, business partnerships, and proprietary data, resulting in higher accuracy and quality. We recommend that the research community select data sources tailored to their specific tasks. Besides, our selection of contextual features is primarily influenced by research in the past few years, which may introduce feature selection bias. For instance, if a contextual feature significantly impacts crowd mobility but hasn't been well studied, it might be overlooked in \texttt{STContext}. To address this issue, we plan to continuously maintain and update \texttt{STContext}, ensuring it remains beneficial for the research community over the long term.

\subsubsection{Future Work}
While we conduct experiments on five typical STCFP tasks, expanding our scope to include a wider range of datasets and tasks may enhance the comprehensiveness of our findings. Moreover, we select MTGNN \cite{MTGNN_2020} as the backbone network for crowd flow prediction due to its strong performance in recent benchmarks \cite{dl_traffic_2021}. However, the variability among different deep STCFP models in capturing spatio-temporal dependencies may not be fully explored, potentially limiting the applicability of our findings across models. Therefore, we will evaluate more advanced backbone networks to improve the generalization of our results.
We also aim to develop a contextual feature selection mechanism that identifies beneficial features for specific tasks. Given our finding that contextual features can degrade performance in context-unrelated scenarios, such a mechanism would help researchers choose effective contexts and exclude irrelevant influences.
Another promising avenue for future work is to enhance data quality through techniques like spatio-temporal imputation. Our experiments indicate (see Section~\ref{sec: real_past_comparison}) that accurate weather forecasts lead to better predictions. Higher-quality contextual features would undoubtedly benefit most context-aware STCFP models.

% \section{Dataset and Code Availability}
% The entire implementation of this work, including the dataset and experimental code, is publicly available at \url{https://anonymous.4open.science/r/STContext-F68A}. The code is released under the MIT License: \url{https://opensource.org/licenses/MIT}. To enhance usability, we also provide detailed metadata descriptions, along with tutorials on loading and using various types of contextual data. Additional details can be found on the project website.

\section{Conclusion}
In this paper, we introduce the \texttt{STContext} dataset, which provides diverse contextual data, including weather, air quality index, holidays, temporal position, points of interest, road networks, demographics, administrative divisions, and spatial position across five STCFP tasks. 
To build this dataset, we focused on three key efforts: first, reviewing contextual features from reputable venues to identify commonly used and publicly accessible data; second, investigating these public data to facilitate data collection from multiple sources; and third, creating a taxonomy that classifies contextual features into spatial, temporal, and spatio-temporal contexts, reflecting their characteristics and guiding effective modeling methods.
To our knowledge, \texttt{STContext} is the most comprehensive dataset currently available for developing context-aware STCFP models. We also propose a unified workflow for incorporating contextual features into deep STCFP models, conducting extensive experiments that offer several findings and guidelines for building effective context-aware STCFP models. We hope \texttt{STContext} will serve as a valuable resource for the STCFP research community, advancing generalizable context modeling techniques.

% \begin{acks}
% \end{acks}

\bibliographystyle{ACM-Reference-Format}
\bibliography{main}

\end{document}